\documentclass[conference]{IEEEtran}
\IEEEoverridecommandlockouts
\usepackage{cite}
\usepackage{amsmath,amssymb,amsfonts}
\usepackage{algorithmic}
\usepackage{graphicx}
\usepackage{textcomp}
\usepackage{xcolor}
\usepackage{amsmath,amssymb,amsfonts}
\usepackage{algorithmic}
\usepackage{graphicx}
\usepackage{textcomp}
\usepackage{multirow}
\usepackage{blindtext, xcolor}
\usepackage{comment}
\usepackage{url}
\usepackage[linesnumbered,ruled]{algorithm2e}
\def\BibTeX{{\rm B\kern-.05em{\sc i\kern-.025em b}\kern-.08em
    T\kern-.1667em\lower.7ex\hbox{E}\kern-.125emX}}
\newcommand{\beginsupplement}{%
        \setcounter{table}{0}
        \renewcommand{\thetable}{S\arabic{table}}%
        \setcounter{figure}{0}
        \renewcommand{\thefigure}{S\arabic{figure}}%
        \setcounter{equation}{0}
        \renewcommand{\theequation}{S\arabic{equation}}%
        \setcounter{section}{0}
        \renewcommand{\thesection}{S\arabic{section}}%
     }
\def\BibTeX{{\rm B\kern-.05em{\sc i\kern-.025em b}\kern-.08em
    T\kern-.1667em\lower.7ex\hbox{E}\kern-.125emX}}
\begin{document}

\title{DAM: Diffusion Activation Maximization for 3D Global Explanations}

\author{\IEEEauthorblockN{Hanxiao Tan}
\IEEEauthorblockA{\textit{AI Group, TU Dortmund} \\
Dortmund, Germany \\
hanxiao.tan@tu-dortmund.de}
}

\maketitle

\begin{abstract}
In recent years, the performance of point cloud models has been rapidly improved. However, due to the limited amount of relevant explainability studies, the unreliability and opacity of these black-box models may lead to potential risks in applications where human lives are at stake, e.g. autonomous driving or healthcare. This work proposes a DDPM-based point cloud global explainability method (DAM) that leverages Point Diffusion Transformer (PDT), a novel point-wise symmetric model, with dual-classifier guidance to generate high-quality global explanations. In addition, an adapted path gradient integration method for DAM is proposed, which not only provides a global overview of the saliency maps for point cloud categories, but also sheds light on how the attributions of the explanations vary during the generation process. Extensive experiments indicate that our method outperforms existing ones in terms of perceptibility, representativeness, and diversity, with a significant reduction in generation time. Our code is available at: \url{https://github.com/Explain3D/DAM}.
\end{abstract}
\section{Introduction}
As AI applications expand into a wide range of industries, the reliability of deep learning (DL) models is becoming as critical as their prediction accuracy. However, most DL models are incomprehensible due to their sophisticated structures, which are known as black-box models. Explainability studies are a potential solution to strengthen model reliability which focuses on providing a better understanding of opaque black-box models by demonstrating decision basis or generalizing model behaviors. The vast majority of existing explainability methods are devoted to investigating tabular \cite{burkart2021survey,das2020opportunities}, image \cite{van2022explainable,nazir2023survey} and text models \cite{cambria2023survey,danilevsky2020survey}.

Despite the extensive explainability studies in the aforementioned fields, relevant researches are still in the preliminary stage for other data formats, such as point clouds. The investigation of the trustworthiness for point cloud models is significant as they are widely applied in robotics \cite{pomerleau2015review}, healthcare \cite{cheng2020morphing} and traffic systems \cite{cui2021deep}, where human life is at stake. Furthermore, explainability methods for images may not be directly transferable to point clouds due to architectural distinctions \cite{tan2023visualizing}. Because of the disordered nature, point cloud models incorporate special structures as alternatives to the 2D convolution kernel, such as global pooling layers \cite{qi2017pointnet} and k-NN local feature extractors \cite{qi2017pointnet++}. Therefore, investigating the explainability of point cloud models may yield novel inspirations in model design or revision.

\begin{figure}
    \begin{centering}
    \includegraphics[width=0.475\textwidth]{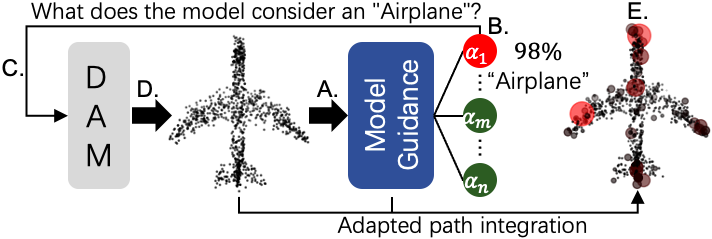}
    \caption{Overview of DAM. A. Feeding the explanation into the classifier. B. Obtaining the target activation value and gradients. C. Guiding the DDPM model with the gradients. D. Generating a better explanation and E. Global saliency map.}
    \label{Fig:overview}
    \end{centering}
\end{figure}

With the above motivation, this work proposes a novel method for visualizing global explanations of point cloud models, named Diffusion Activation Maximization (DAM). DAM is based on a neural network visualisation technique called \textit{Activation Maximization} \cite{erhan2009visualizing} (AM), and learns the distribution of the dataset through a diffusion model to yield high-quality global explanations. An overview is shown in Fig. \ref{Fig:overview}. Moreover, incorporating and adapting the path attribution method \textit{Path Integrated Gradients} \cite{sundararajan2017axiomatic}, we propose a plausible global saliency map generating method adaptable to DAM, called Integrated Gradients for Diffusion (IGD), which not only provide global attributions for point cloud categories, but also exhibit variations of critical attributions in the generation process. Compared with existing approaches, our method outperforms both in qualitative and quantitative evaluations. In summary, our contributions are as follows:

\begin{itemize}
    \item We propose a novel global explanation generating method for point clouds based on Denoising Diffusion Probabilistic Models (DDPM), which outperforms existing approaches in terms of performance and speed. To the best of our knowledge, this is the first work incorporating DDPM models for generating global explanations.
    
    \item We adapt the path-based attribution method to DAM, which provides global saliency maps for point cloud categories, while demonstrating how attribution varies during the explanation generating process, thus rendering the explanations more intuitive and interpretable.
\end{itemize}

The paper is structured as follows: The related work is introduced in Sec. \ref{RelatedWork}, the algorithmic background and the proposed approach are elaborated in Sec. \ref{Methods}, the experiments are shown in Sec. \ref{sec:experiments}, and Sec. \ref{sec:conclusion} concludes with a summary.
\section{Related Work} \label{RelatedWork}

In this section we outline the relevant domains and existing research concerning this work.

\textbf{Explainability methods: } Explainability methods can be mainly categorized into two groups, namely, local and global approaches. Local approaches \cite{li2020learning} are dedicated to providing explanations for individual decisions, whose expressions are in the form of saliency maps \cite{simonyan2013deep,bach2015pixel,selvaraju2017grad,springenberg2014striving,sundararajan2017axiomatic,ribeiro2016should,lundberg2017unified,petsiuk2018rise} and counterfactuals \cite{verma2020counterfactual,mothilal2020explaining,goyal2019counterfactual,wachter2017counterfactual}. The former exhibits which features in the input are essential to the decision, while the latter provides an example that is analogous to the input but is predicted to be another class. Global methods present a holistic summary for the entire model and dataset. For models with simpler structures, approximated decision rules can be generalized based on the input-output dependencies \cite{sushil2018rule,hailesilassie2016rule}. For models with complex structures or high-dimensional inputs, rule induction is difficult to accurately cover all decision logics, and a more common technique is to visualize category-representative input examples by highly activating specific neurons through back-propagation, which is known as AM \cite{erhan2009visualizing,nguyen2015deep,simonyan2013deep,nguyen2016synthesizing,nguyen2017plug}. 

AM was first proposed by \cite{erhan2009visualizing}. A straightforward AM application to neural networks fails to produce perceptible images \cite{nguyen2015deep}. Subsequent studies suggested that incorporating constraints or priors enhances the perceptibility of the explanations, e.g. performing L2-norm \cite{simonyan2013deep}, Total Variation \cite{mahendran2016visualizing} and Gaussian blur \cite{yosinski2015understanding} on gradients, or starting optimization from the average of the dataset \cite{mordvintsev2015inceptionism}. Higher quality AM explanations are achieved with the introduction of generative models. Alternately adding the gradients of GANs or Autoencoders during optimization effectively guides the AM explanations closer to real images  \cite{nguyen2016synthesizing,zhou2017activation,nguyen2017plug}.

\textbf{Point cloud models: }The neural network handling raw point clouds was first proposed by \cite{qi2017pointnet}, which achieves a remarkable accuracy by extracting global features through a max-pooling layer. Subsequent refinement includes: extraction of local relative features \cite{qi2017pointnet++,lan2019modeling,liu2019relation}, redefinition of convolution \cite{komarichev2019cnn}, and introduction of tree structures \cite{riegler2017octnet,klokov2017escape,zeng20183dcontextnet} and graphs \cite{simonovsky2017dynamic,wang2019dynamic,chen2019clusternet}.

\textbf{Explainability for point clouds: }Compared to images, the explainability of point cloud models has not been adequately addressed. Several studies demonstrate the local saliency maps of the inputs by perturbation \cite{zheng2019pointcloud} or transplanting explainability methods from images \cite{gupta20203d,tan2022surrogate}. \cite{tan2023visualizing} is the sole study that investigates the global explainability of point cloud models, where they constrain the optimization process with Autoencoders in various architectures to yield perceptible explanations. Moreover, they propose a quantitative evaluation metric specifically for point cloud AM generations. In comparison, we exploit diffusion models that have been shown to beat Autoencoders in image generation \cite{dhariwal2021diffusion} to guide the optimization path. Our method perform superior both qualitatively and quantitatively and can be combined with path-integration approach \cite{sundararajan2017axiomatic} to illustrate global point-wise attributions during optimization. 
\section{Methods} \label{Methods}

In this section we first introduce the algorithmic background (Sec. \ref{preliminaries}), and then elaborate on the proposed method DAM (Sec. \ref{DAM}). Finally, we detail IGD, the adapted path integration approach for DAM (Sec. \ref{Sec:IGD}).

\subsection{Preliminaries} \label{preliminaries}
Consider a point cloud dataset $X=\lbrace x_1,...x_m \mid  x_i \in \mathbb{R}^{N\times D}\rbrace$, where $N$ and $D$ denotes the number of points composing an instance and its dimension, respectively. A well-trained classifier $F$ is the model to be explained, which can be formulated as $F:\mathbb{R}^{N \times D} \mapsto [0,1]^{1 \times N_C}$.

\textbf{Activation Maxmization (AM): } Generally, AM is formulated as:

\begin{equation} \label{eq:am}
x^* = \underset{x}{\mathrm{argmax}}\, (a_i^l(F,x))
\end{equation}
where $a_i^l$ is the $i^{th}$ neuron of layer $l$ on $F$. Typically, the neuron $F_o^i\in \mathbb{R}^{1 \times 1}$ is chosen which is the $i^{th}$ neuron on the logits layer and indicates a particular category. Due to the intrinsic distribution of the data, optimization by AM alone cannot generate perceptible explanations \cite{nguyen2015deep}. External constraints are required which force the optimization towards the path that is compatible with the data distribution. However, \cite{tan2023visualizing} demonstrates that adding ordinary algebraic restrictions on the gradients (e.g., L2-norm \cite{simonyan2013deep}, Total Variation \cite{mahendran2016visualizing}, or Gaussian blur \cite{yosinski2015understanding}) are of limited use for point clouds. They employ Autoencoders to filter out samples from the $R^{N\times D}$ space that are perceivable and then optimize them with AM. Specifically, the explanations can be represented as:

\begin{equation}
    x^*\sim p_{G}(x)p_{F_i}(y|x)
\end{equation}
where $x^*$ fulfills the distribution that both highly activates a neuron ($p_{F_i}(x|y)$, $y$ is the label) and resembles real examples ($p_{G}(x)$), and thus is perceptible .

\textbf{Denoising Diffusion Probabilistic Model (DDPM): } DDPM was first proposed by \cite{sohl2015deep,ho2020denoising}, which is a denoising generative model based on Markov chains. The diffusion model is composed of two phases, with the forward being the diffusion phase, where Gaussian noise is gradually added from a real instance $x_0$:

\begin{equation} \label{DDPMtrain}
    q(x_{1:T}|x_0)=\prod_a^bq(x_t|x_{t-1})
\end{equation}
with the kernel:
\begin{equation} \label{DDPMtrainKernel}
    q(x_t|x_{t-1})=\mathcal{N}(x_t;\sqrt{1-\beta_t}x_{t-1},\beta_t\mathbf{I})
\end{equation}
where $\beta_t$ denotes the variance schedule at step $t$. The reverse is the sampling phase, which starts from a Gaussian distributed noise $x_T=\mathcal{N}(0,\mathbf{I})$:

\begin{equation} \label{DDPMsample}
    p_\theta(x_{0:T})=p(x_T)\prod_{t=1}^Tp_\theta(x_{t-1}|x_t)
\end{equation}
with
\begin{equation} \label{DDPMsampleKernen}
    p_\theta(x_{t-1}|x_t)=\mathcal{N}(x_{t-1};\mu_\theta(x_t,t),\Sigma_\theta(x_t,t))
\end{equation}
where $\mu_\theta$ and $\Sigma_\theta$ are estimated by the model with parameter $\theta$. There are numerous subsequent improvements to DDPM \cite{dhariwal2021diffusion}, including point cloud applicable ones \cite{Luo_2021_CVPR,lyu2021conditional,nichol2022point,li2022diffusionpointlabel}.

\textbf{Path Integrated Gradients}: Path Integrated Gradients \cite{sundararajan2017axiomatic} is a series of gradient-based explainability methods. By determining an uninformative baseline and a path $\gamma:[0,1]\rightarrow R^{N \times D}$, the gradient from the baseline to the input is accumulated along the path in order to observe which are the critical features for the prediction. The general form of Path Integrated Gradients is: 

\begin{equation} \label{IGformular}
    PathIG_i^\gamma(x)=\int_{\alpha=0}^1\frac{\partial F(\gamma(\alpha))}{\partial \gamma(\alpha)}\frac{\partial \gamma(\alpha)}{\partial \alpha}d\alpha
\end{equation}
where $\alpha=0$ and $1$ indicate the baseline and input, respectively.

\subsection{Diffusion Activation Maximization (DAM)} \label{DAM}

\begin{figure*}
    \begin{centering}
    \includegraphics[width=0.75\textwidth]{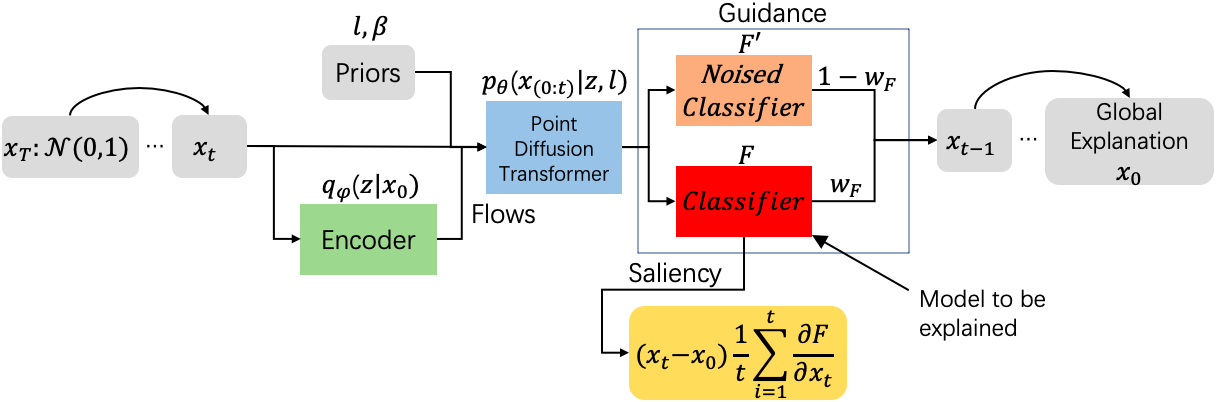}
    \caption{Overview of the DAM structure. There are two main explanations, one for the globally explainable sample $x_0$ (gray block on the right), and the other for the saliency map of the diffusion process (yellow block below). }
    \label{Fig:PDTStruc}
    \end{centering}
\end{figure*}

DAM consists of two components, generative training and explanation sampling. An overview of the structure can be seen in Fig. \ref{Fig:PDTStruc}.

\textbf{Generative training: } We leverage DDPM (equation \ref{DDPMtrain} to \ref{DDPMsampleKernen}) to filter the perceptible $p_G(x) \sim X$ from $\mathbb{R}^{N \times D}$. Compared to \cite{tan2023visualizing} which utilizes Autoencoders, the advantages of DDPM are twofold:

\begin{itemize}
    \item Existing study show that DDPM possesses greater potential for image synthesis compared to other generative models \cite{dhariwal2021diffusion}, which may be applicable for point clouds as well.

    \item DDPM handles noise for each point independently, which is ideally suited to the disorderly nature of point clouds. This property enables the follow-up research such as critical attribution analysis.
\end{itemize}

The training process is roughly analogous to image DDPM, and we follow and adapt \cite{Luo_2021_CVPR}, exploiting the following objective as the training loss for point cloud DDPM

\begin{equation}
\begin{aligned}
L(\theta,\varphi,\alpha)=&\mathbb{E}_q \left[ \right. D_{KL}(q_{\varphi}(z|x_0)\||p_\omega (\omega )\cdot \left |det\frac{\partial F_\alpha }{\partial \omega }  \right |^{-1}) \\
+ & \sum_{t>1}\sum_{i=1}^n D_{KL}(q(x_{t-1}|x_t,x_0)||p_\theta(x_{t-1}|x_t,z,l)) \\
-&\sum_{i=1}^n log(p_\theta(x_0|x_1,z,l))\left. \right]
\end{aligned}
\end{equation}
where $x_t$ denotes the input at time point $t$, $D_{KL}$ indicates Kullback–Leibler divergence and $q_{\varphi}(z|x_0)$ is a variant of PointNet \cite{qi2017pointnet} that serves as an encoder whose output is the mean and variance of $x_0$. $p_\omega (\omega )\cdot \left |det\frac{\partial F_\alpha }{\partial \omega }  \right |^{-1}$ are affine coupling layers, which project isotropic Gaussian distributions $p_\omega (\omega )$ onto more complex distributions via a trainable bijective $F_\alpha$ as the training priors \cite{Luo_2021_CVPR}.

To enhance performance and interpretability, we incorporate the following improvements:

\begin{itemize}
    \item Following \cite{nichol2021improved}, to prevent $x_t$ from collapsing into pure noise at a small $t$, the noise schedule $\alpha_t$ is optimized as
    \begin{equation}
        \alpha_t=\frac{f(t)}{f(0)},f(t)=cos(\frac{t/T+\beta_1}{1+\beta_1}\cdot \frac{\pi}{2})^2
    \end{equation}

    \item Label $l$ is embedded in the training process. Class information guides the sampling better towards a specific category. We simply convert the labels into one-hot vectors and concatenate them with the coordinate data.
    
    \item We propose a novel model: Point Diffusion Transformer (PDT). Transformers are shown to be powerful architectures for learning latent representatives, which has already been introduced into point cloud DDPM \cite{nichol2022point}. However, we argue that the existing models (including non-Transformers as \cite{Luo_2021_CVPR}) neglect the ``symmetry property" , which was first proposed by \cite{qi2017pointnet}, that the output of a point cloud model should be independent of the inputs sequence. Especially with DDPM, the noise added is typically isometric \cite{ho2020denoising}. Thus, the utilization of asymmetric components such as multi-size convolutional kernels or fully connected layers for non-global features should be minimized. Moreover, eliminating correlations between points results in cleaner gradients in AM iterations and hence reinforces the representativeness of explanations.

    To address the above objective, we concatenate the point-wise coordinates, priors and label vector and obtain the input $x^*\in \mathbb{R}^{N \times (D+D_p+D_l)}$ ($D_p$ and $D_l$ are the dimensions of the prior and label vectors, respectively, see Fig. \ref{PDT_general} c)). It is subsequently fed into a Point Diffusion Encoder (PDE), which is a multi-headed self-attention module whose inputs of Query, Key and Value are $x^*$. The following module is a Point Diffusion Decoder (PDD), which shares a similar structure to PDE, except that the inputs of Query and Key are the residuals $x^*+PDE(x^*)$ while the input Value is $x^*$ (Fig. \ref{PDT_general} b)). Further discussions for the input of the attention headers can be found in Sec. \ref{sec:qkvstudy}). The most crucial property of PDT is that the entire model utilizes only the convolution kernel of $D_{in} \times 1$, ensuring that each point is independent of the input order during noise prediction and AM optimization (see Sec. \ref{sup:featindependent} for ablation study). This structure not only alleviates possible gradient conflicts between AM and diffusion processes, but also eliminates interference when calculating gradient integration, resulting in cleaner saliency maps (see Sec. \ref{Sec:IGD}). The detailed architecture of PDT is shown in Fig. \ref{PDT_general} a).
    
\end{itemize}

\textbf{Explanation sampling} consists of two parts, sampling and explaining processes, corresponding to perceivability and representativeness of the explanations, respectively.

The sampling process is a reversed denoising Markov chain where the Gaussian noise $X_t$ is fed into the trained diffusion model $p_\theta$, and a perceptible synthetic sample $x_0$ is obtained after multi-step denoising. The reverse diffusion process can be formulated as

\begin{equation}\label{equa:diffusion}
    p_\theta(x_{(0:t)}|z,l)=p(x_t)\prod_{t=0}^{t}\mathcal{N}(x_{t-1}|\mu_\theta(x_t,t,z,l),\beta_t\mathbf{I})
\end{equation} 

Practically, we randomize an $x_{r}$ and input it into $q_{\varphi}(z|x_0)$ to obtain encoded flow $z_{\mu}$ and $z_{\sigma}$, then reparameterize them to $z$, and generate $x_{T}$ via equation \ref{equa:diffusion}. The advantage of this initialization is that it employs $q_{\varphi}(z|x_0)$, which is well-trained to normalize $x_T$ to better approximate $z$ of real data. Detailed analysis is provided in Sec. \ref{sec_random_x_z}.

In the explaining phase, following Equation \ref{eq:am}, we force the reversed diffusion process in the direction that highly activates a certain neuron $a_i^l$ of the classifier $F$ by implanting a guidance gradient $x_t = x_t + \frac{\partial \alpha_i^l}{\partial x_t}$. The challenge of incorporating DDPM model is that, as $t$ approaches $T$, the diffusion samples approximate pure Gaussian distributions $x_t\rightarrow \mathcal{N}(0,\mathbf{I})$, $F$ may never see analogous inputs and the obtained guidance gradients may be biased. Inspired by \cite{dhariwal2021diffusion}, we enable a twin classifier $F'$ trained on a noised dataset $X'$ as a transition. $F'$ shares the identical architecture as $F$, and is acquired by continuing training on $X'$ after $F$ converges on $X$, where $X'$ is the noisy version of $X$, which is transformed from the forward diffusion process $q(x_{1:t}|x_0)$. It contains samples of various noise levels based on the time information $t$ (technical details are provided in Sec. \ref{sec:traindetail}). We train $F'$ by fusing binarized vector $t$ with the coordinates information of $X'$ so that it better guides the diffusion gradients. As $t$ approaches $0$, the sample outline is gradually regularized and the guiding classifier needs to be switched to $F$ (the model to be explained). We schedule two weights $W_F$ and $W_{F'}$ such that $W_F + W_{F'} = 1$, which weight the guidance gradients of $F$ and $F'$, respectively: $W_{F'}$ converges to $1$ as $t$ approaches $T$, while $W_F = 1$ when $t = 0$ (analytical comparisons can be found in Sec. \ref{subsection:eval_noised}). During optimization, we choose $log(SoftMax)$ as the target activation $\alpha$, which significantly enhances the explanation performance (refer to Sec. \ref{sec:logitsvssoftmax} for further comparisons). A general overview of our sampling approach can be found in Algorithm \ref{algo:sample}. 

\begin{algorithm} 
    \SetKwInOut{Input}{Input}
    \SetKwInOut{Output}{Output}
    
    \Input{Class label $l$, guidance weights $W_F$ and guidance scale $s$}
    \Output{Global explanation $x_0$}
    $Sample\ x_t\sim \mathcal{N}(0,\mathbf{I})$\;
    \textbf{for all} $t$ from $T$ to $1$ \textbf{do} \\
    {
        \hskip1.5em$\mu_t$,$\sum_t$ $\leftarrow$ $(\mu_\theta(x_t,l),\Sigma_\theta(x_t,l))$
        
        \hskip1.5em$Sample \ x_{t-1}\sim (\mathcal{N}(\mu_t+s\Sigma_t(W_F\bigtriangledown_{x_t}log(F(x,t))+(1-W_{F})\bigtriangledown_{x_t}log(F'(x,t))),\Sigma_t))$
    }
    
    \textbf{end for}

    \textbf{return} $x_0$

    \caption{Sampling algorithm of DAM, given a diffusion model $(\mu_\theta(x_t,l),\Sigma_\theta(x_t,l))$, a noised classifier $F'(x,t)$ and the model to be explained $F(x)$}\label{algo:sample}
\end{algorithm}

\subsection{Integrated Gradients for Diffusion (IGD)} \label{Sec:IGD}

For DAM, we propose an improved path method Integrated Gradients for Diffusion (IGD), which exhibits more plausible global saliency maps for point clouds and depict attribution variations in the diffusion process. IGD can be formulated as 

\begin{equation}
    IGD=(x_0-x_T)\times\sum_{t=0}^{T}\frac{\partial F(p_\theta(x_{(0:t)}) }{\partial x_t} \times \frac{1}{T}
\end{equation}

The two elements of Path Integrated Gradients are the baseline and the path, respectively \cite{sundararajan2017axiomatic}, which we adapt as

\begin{itemize}
    \item \textbf{Baseline: } The baseline is defined as ``uninformative" \cite{sturmfels2020visualizing}, which ensures that the integrated gradient captures the whole attributional variation of the model. We consider $x_T$ as baseline as it is sampled from $\mathcal{N}(0,\mathbf{I})$ and does not contain any information.
    \item \textbf{Path: } Various options are available from the baseline to the input, with \textit{linear paths} ($x'+\alpha \times(x-x')$, $\alpha \in [0,1]$) being the frequent option. In IGD, we leverage the diffusion sampling process itself $p_\theta(x_{(0:t)})$ as the path.
\end{itemize}

In the sampling process, we have calculated exactly the guide gradients of $F$: $\frac{\partial F(p_\theta(x_{(0:t)}) }{\partial x_t}$, thus there is no need to recalculate the gradients and simply integrate them from $x_T$ to $x_0$. The procedure of IGD is described in Algorithm \ref{algo:tracking}. 

\begin{algorithm} 
    \SetKwInOut{Input}{Input}
    \SetKwInOut{Output}{Output}

    \Output{Saliency maps $IGD$ for arbitrary time step $t$}
    $\Delta_g = 0$
    
    \textbf{for all} $t$ from $T$ to $1$ \textbf{do} \#Sampling starts \\
    {
        \hskip1.5em$\Delta_g \leftarrow \Delta_g + \frac{\partial F(x_t)}{\partial x_t}$

        \hskip1.5em$IGD \leftarrow (x_t-x_T) \times \Delta_g \times{\frac{1}{T-t}}$ 

        \hskip1.5em Output(IGD)  \#IGD can output in arbitrary loop
        
        \hskip1.5em$x_{t-1} \leftarrow p_\theta(x_t)$   \#Then perform DAM Sampling process
        
    }
    
    \textbf{end for}

    \caption{Integrated Gradients for DAM (IGD), given a DAM model $p_\theta(x_{0:T})$ and the model to be explained $F(x)$}\label{algo:tracking}
\end{algorithm}

\begin{figure}
    \begin{centering}
    \includegraphics[width=0.3\textwidth]{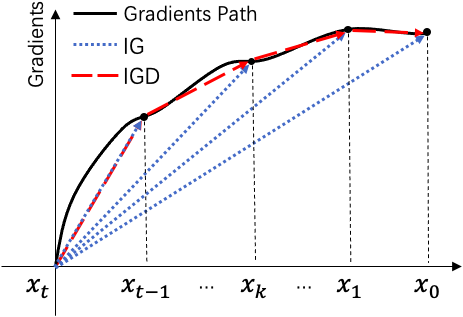}
    \caption{Visual comparison of path methods IGD and typical IG. The gradient path in the diffusion process is the integration from $x_0$ to $x_T$ (the black curve). Typical IG paths for $x_t$ are the linear integration of $x_0$ to $x_t$, which may lead to bias, while the path of IGD for $x_t$ is the integration from $x_{t-1}$ to $x_t$, which better approximates the real path.}
    \label{Fig:IGvsIGD}
    \end{centering}
\end{figure}

Compared to standard IG, IGD is more flexible for gradients integration in diffusion processes. As shown in Fig. \ref{Fig:IGvsIGD}, assume that the path of the sample gradients from a diffusion process is given by the black curve. For all $x_t$, the typical IG integrates the linear path of the gradients starting from the baseline each time (the blue line), which results in a bias between the final integration and the real one. This bias causes unfaithfulness and large fluctuations in the generated saliency maps. In comparison, the issue is significantly alleviated by IGD. IGD integrates the gradients of $x_{t-1}$ and $x_t$ with a linear path, which minimizes the bias under the precondition that the true gradient path is unavailable. We quantitatively compare the performance of the two path methods in terms of coherence and sensitivity (faithfulness) in Sec. \ref{subsection:eval_IGD}. From an explainability perspective, IGD offers two advantages: a) It provides inductive exhibitions of feature attributions from a global perspective (rather than local specific inputs), b) The high confidence of examples ensured by AM enables the attributions to be globally representative of the corresponding categories.

\section{Experiments} \label{sec:experiments}
In this section we present the qualitative demonstrations (Sec. \ref{subsection:qual}) and quantitative evaluations (Sec. \ref{subsection:quan}) for DAM, and visualization of IGD and the corresponding quantitative assessments (Sec. \ref{subsection:eval_IGD}).

We employ \textit{ModelNet40} \cite{wu20153d} as the primary experimental dataset (default dataset, unless specifically mentioned), it contains $12311$ CAD models, of which $9843$ are used for training and $2468$ for testing. In addition, we validate the performance of our method on ShapeNet, a larger database of 3D objects containing a total number of $45969$ samples in $55$ categories, of which $35708$ are used for training and $10261$ for test. During the sampling phase, we generate 10 samples containing $1024$ points for each class and randomly select $5$ from the real dataset as the benchmark for calculating the Chamfer and Fréchet inception distances (CD and FID) for quantitative evaluations. Detailed model training configurations can be found in Sec. \ref{sec:traindetail}. In qualitative and quantitative comparisons, we take Autoencoder-based AE, AED, and NAED \cite{tan2023visualizing} as the main competitors, which are currently the sole existing global explainability methods for point clouds based on Activation Maximization.

\subsection{Qualitative Visualizations for DAM} \label{subsection:qual}
In this section, we qualitatively demonstrate the visualization of the global explanations generated by DAM, including perceptibility and diversity. 

\begin{figure}
    \begin{centering}
    \includegraphics[width=0.475\textwidth]{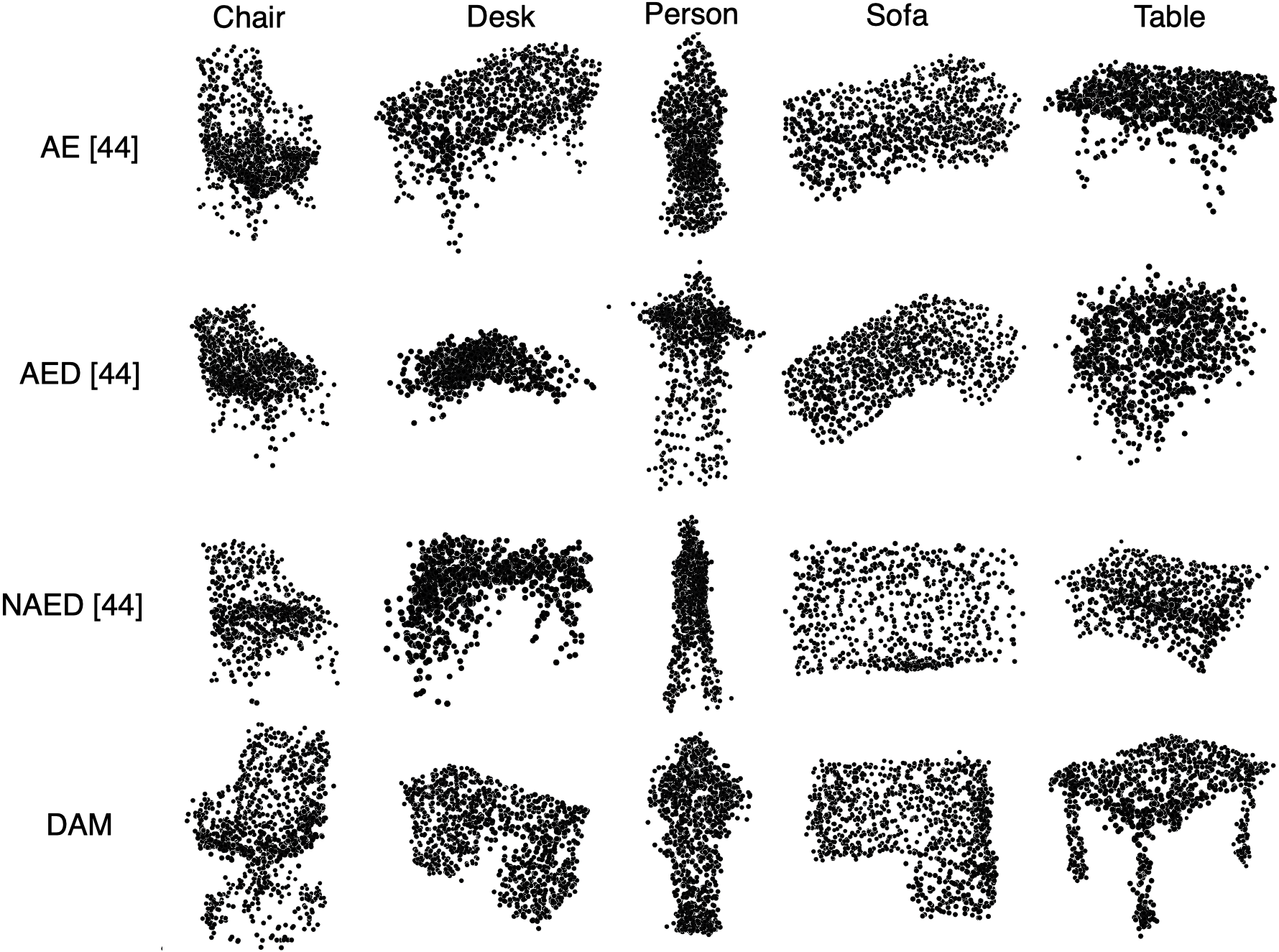}
    \caption{Global explanations of 5 classes generated by DAM. For comparison, we present the identical amount of explanations generated by AE, AED and NAED \cite{tan2023visualizing}. More visualizations are shown in Fig. \ref{More_DAM}.}
    \label{Fig:percept}
    \end{centering}
\end{figure}

\textbf{Perceptibility} is the degree to which the generated explanation can be comprehended by humans. Generally, complete and high-quality explanations are more perceptible. We select common classes from the 40 categories of ModelNet40 and generate global explanations with DAM (illustrated in Figure \ref{Fig:percept}). We also qualitatively compare the results of DAM with AE, AED and NAED \cite{tan2023visualizing}. Overall, the geometric structure of DAM-generated explanations is more robust and thus more easily perceived by humans. Besides the output layer, an interesting observation is to visualize the middle layer of the network, and we demonstrate part of the results in Sec. \ref{sup:visu_other_layers}.

\begin{figure}
    \begin{centering}
    \includegraphics[width=0.4\textwidth]{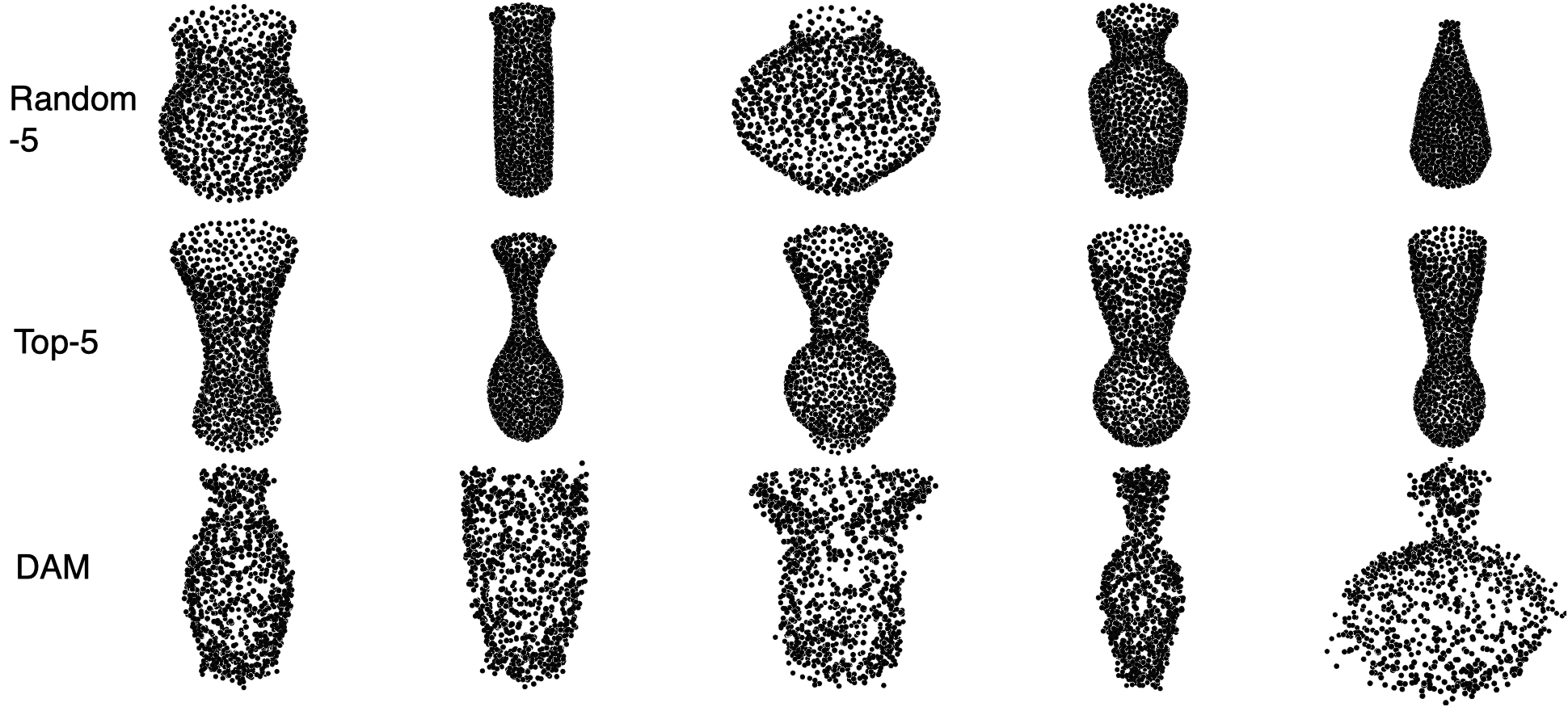}
    \caption{Diversity examples. We randomly generated 5 explanations for the category ``vase". For intuition, we also show 5 randomly chosen objects of the same class from the dataset (Random-5), and 5 samples that most highly activate the neuron ``vase" (Top-5). More diversity is displayed in Fig. \ref{More_DAM}.}
    \label{Fig:Diversity}
    \end{centering}
\end{figure}

\textbf{Diversity} is also one of the essential properties for explanations. Abundant diversity provides humans with different perspectives of explanations to gain better comprehension \cite{nguyen2017plug}. Fig. \ref{Fig:Diversity} illustrates the diversity of explanations generated by DAM. It can be observed that DAM is able to generate diversified and qualified global explanations.

\begin{figure}
    \begin{centering}
    \includegraphics[width=0.475\textwidth]{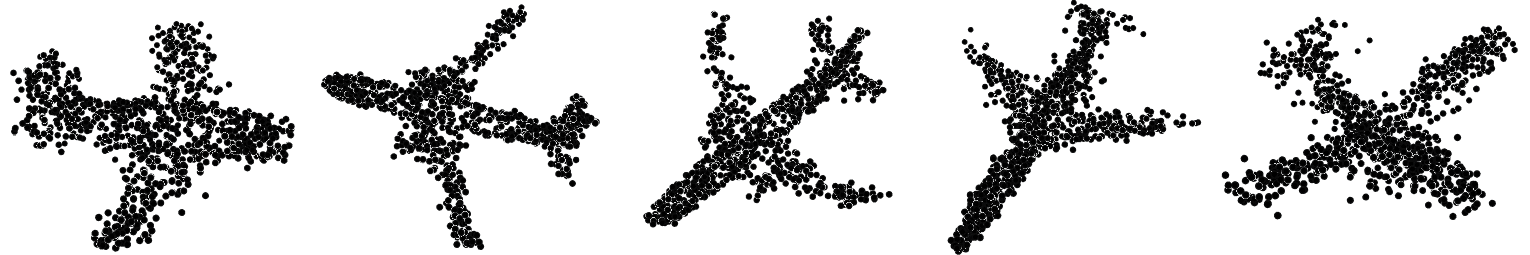}
    \caption{five global explanations generated by DAM on ShapeNet with the category ``Airplane". }
    \label{Fig:ShapeNet}
    \end{centering}
\end{figure}

\textbf{Visulization on ShapeNet:} We exhibit in Fig. \ref{Fig:ShapeNet} five global explanations of the class ``Airplane" from ShapNet. In conclusion, our approach achieves both performance and diversity on different datasets. 

\begin{figure}
    \begin{centering}
    \includegraphics[width=0.35\textwidth]{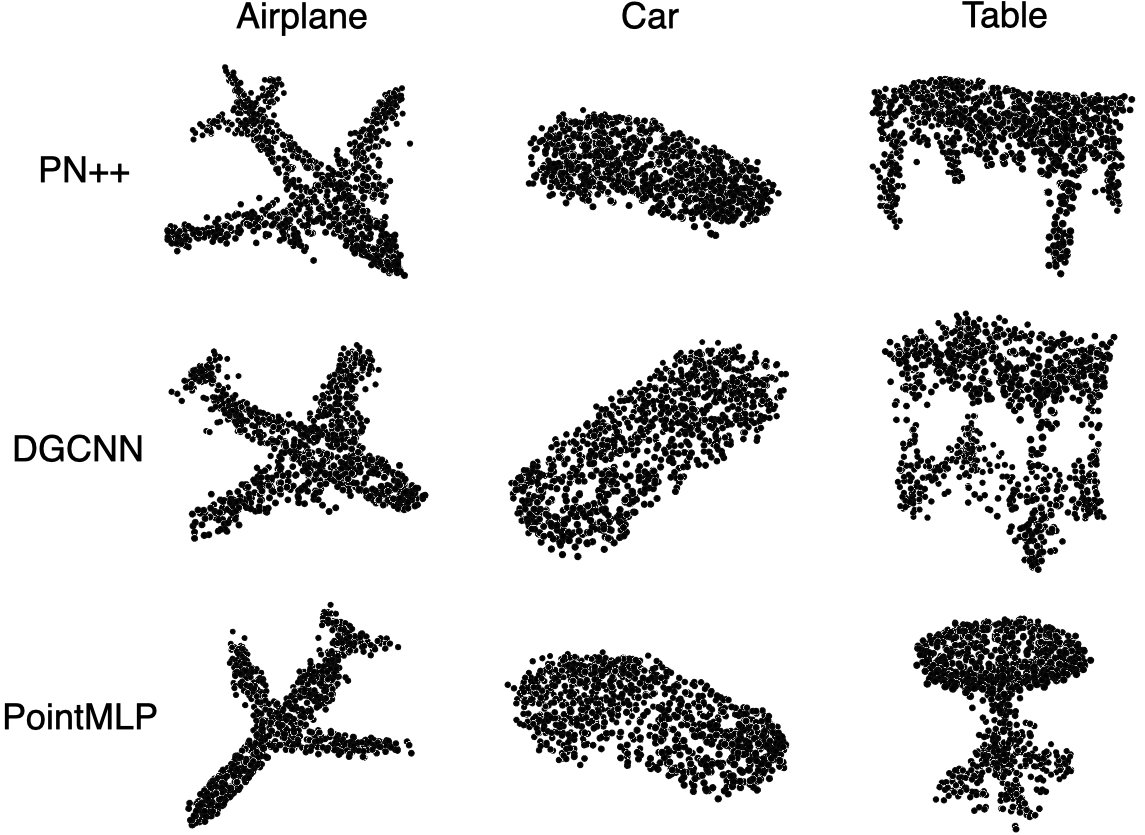}
    \caption{Global explanations of other models generated by DAM. From top to bottom are PointNet++, DGCNN and PointMLP.}
    \label{Fig:othernets}
    \end{centering}
\end{figure}

\textbf{Explanations on other models:} We test DAM on other popular or state-of-the-art point cloud models besides PointNet, including PointNet++ \cite{qi2017pointnet++}, DGCNN \cite{wang2019dynamic}, and PointMLP \cite{ma2022rethinking}. Fig. \ref{Fig:othernets} depicts the visualization of their global explanations generated by DAM. It can be observed that the performance of DAM in explaining models is independent of their internal architectures.

\subsection{Quantitative Evaluation for DAM} \label{subsection:quan}

\begin{table*}[]
\centering
\begin{tabular}{ccc|cccc|c}
\hline
Dataset                                                                              & Method              & Model                          & m-IS $\uparrow$                 & FID $\downarrow$                & CD $\downarrow$                 & EMD $\downarrow$                  & PCAMS $\uparrow$              \\ \hline
\multirow{4}{*}{\begin{tabular}[c]{@{}c@{}}ModelNet40 \\ / \\ ShapeNet\end{tabular}} & \multirow{3}{*}{AE} & AE \cite{tan2023visualizing}   & 1.085 / 1.012                   & 0.016 / 0.017                   & \textbf{0.044} / \textbf{0.047} & 143.13 / 147.87                   & 4.71 / 4.57                   \\
                                                                                     &                     & AED \cite{tan2023visualizing}  & 1.124 / 1.146                   & 0.018 / 0.012                   & 0.086 / 0.076                   & 241.35 / 208.02                   & 4.37 / 4.65                   \\
                                                                                     &                     & NAED \cite{tan2023visualizing} & 1.461 / 1.157                   & 0.014 / 0.011                   & 0.074 / 0.067                   & 207.65 / 203.74                   & 4.89 / 4.75                   \\ \cline{2-8} 
                                                                                     & DDPM                & DAM (ours)                     & \textbf{1.781} / \textbf{1.706} & \textbf{0.009} / \textbf{0.010} & 0.045 / 0.054                   & \textbf{133.97} / \textbf{146.21} & \textbf{5.68} / \textbf{5.45} \\ \hline
\end{tabular}
\caption{Quantitative evaluation of the explanations generated by DAM compared with the models proposed in \cite{tan2023visualizing}. As a reference, we additionally introduce Earth Mover's Distance. The up and down arrows denote that higher and lower values indicate better performance, respectively. }
\label{tab:Quan_compare}
\end{table*}

We employ the identical evaluation metrics as in \cite{tan2023visualizing}, whose detailed formulations can be found in Sec. \ref{sup:intrometrics}. We demonstrate the quantitative results of DAM in Tab. \ref{tab:Quan_compare}. For comparison, we also exhibit the assessment of the explanations generated by existing studies: AE, AED and NAED \cite{tan2023visualizing}. The results indicate that DAM outperforms all existing point cloud global explainability methods in each metric, except for sacrificing a minimal point-wise distance to balance the diversity compared with AE. Approximate outcomes are yielded from the evaluation on ShapeNet.

Besides, as the interpolation times in the diffusion process are restricted, the processing time is significantly reduced. Tab. \ref{tab:proc_time} details the average time consumption of generating an explanation, again, we compare with AE, AED and NAED from \cite{tan2023visualizing}. 

Tab. \ref{tab:othermodels} shows the quantitative evaluations of explaining other models generated by DAM. It can be seen that DAM outperforms existing methods \cite{tan2023visualizing} on all metrics except for few ones on PointNet++ where DAM is slightly inferior.

\begin{table}[]
\centering
\resizebox{0.475\textwidth}{17mm}{
\begin{tabular}{ccccccc}
\hline
                       & Model      & m-IS$\uparrow$ & FID$\downarrow$ & CD$\downarrow$ & EMD$\downarrow$ & PCAMS$\uparrow$ \\ \hline
\multirow{4}{*}{PN2}  & AE \cite{tan2023visualizing}         & 1.103           & 0.008            & \textbf{0.041}  & \textbf{134.16}  & 5.12             \\
                       & AED \cite{tan2023visualizing}       & 1.107           & 0.020            & 0.122           & 255.46           & 4.12             \\
                       & NAED \cite{tan2023visualizing}       & \textbf{1.866}  & 0.011            & 0.072           & 236.42           & 5.43    \\
                       & DAM & 1.695           & \textbf{ 0.008}   & 0.048           & 134.41           & \textbf{5.62}            \\ \hline
\multirow{4}{*}{DGC} & AE \cite{tan2023visualizing}         & 1.020           & \textbf{0.010}   & 0.105           & 252.82           & 4.43             \\
                       & AED \cite{tan2023visualizing}        & 1.358           & 0.013            & 0.109           & 343.15           & 4.63             \\
                       & NAED \cite{tan2023visualizing}      & 1.316           & 0.015            & 0.109           & 335.51           & 4.52             \\
                       & DAM & \textbf{1.758}  & \textbf{0.010}   & \textbf{0.047}  & \textbf{130.52}  & \textbf{5.58}   \\ \hline
PML               & DAM & 1.49            & 0.009            & 0.047           & 129.79           & 5.37            \\ \hline
\end{tabular}}
\caption{Quantitative evaluations of global explanations generated by DAM on other point cloud models. In the first column, PN2, DGC and PML indicate the experiment results on PointNet++, DGCNN and PointMLP, respectively.}
\label{tab:othermodels}
\end{table}


\begin{table}[]
\centering
\begin{tabular}{c|cccc}
\hline
        & AE \cite{tan2023visualizing}    & AED \cite{tan2023visualizing}    & NAED \cite{tan2023visualizing}   & DAM            \\ \hline
$\hat{t}$ (s) & 47.75 & 458.69 & 201.27 & \textbf{12.35} \\ \hline
\end{tabular}
\caption{Average time $\hat{t}$ required to generate an explanation. Note that we report the processing time for comparable performance rather than identical number of iterations.}
\label{tab:proc_time}
\end{table}

\subsection{Visualizations and assessments for IGD} \label{subsection:eval_IGD}

\begin{table*}[]
\centering
\begin{tabular}{ccccccc}
\hline
           & \multicolumn{2}{c}{Faithfulness}              & Global Stability               & \multicolumn{3}{c}{Local Continuity}                                        \\ \hline
           & $S^{j=0.5}_F\uparrow$ & $S^{j=1.0}_F\uparrow$ & $L_{var}\downarrow$            & $L_{D}\downarrow$              & $L_{W}\downarrow$ & $L_{SC}\uparrow$       \\ \hline
RDM        & $-0.120$              & $1.064$              & $2.411$                        & $1.087\times 10^{-3}$          & $5464.640$        & $-8.784\times 10^{-4}$ \\
IG         & $1.438$              & $4.226$              & $0.037$                        & $1.989\times 10^{-4}$          & $364.996$         & $\mathbf{0.986}$       \\
IGD (ours) & $\mathbf{38.974}$      & $\mathbf{91.061}$      & $\mathbf{2.630\times 10^{-8}}$ & $\mathbf{1.892\times 10^{-6}}$ & $\mathbf{0.107}$  & $0.753$                \\ \hline
\end{tabular}
\caption{Quantitative evaluation of attributions in diffusion. RDM is a set of randomly generated attributions for reference. IG and IGD are the conventional IG with linear paths and the gradient integration with diffusion paths proposed in this paper, respectively.}
\label{tab:IGD_compare}
\end{table*}

\begin{figure}
    \begin{centering}
    \includegraphics[width=0.475\textwidth]{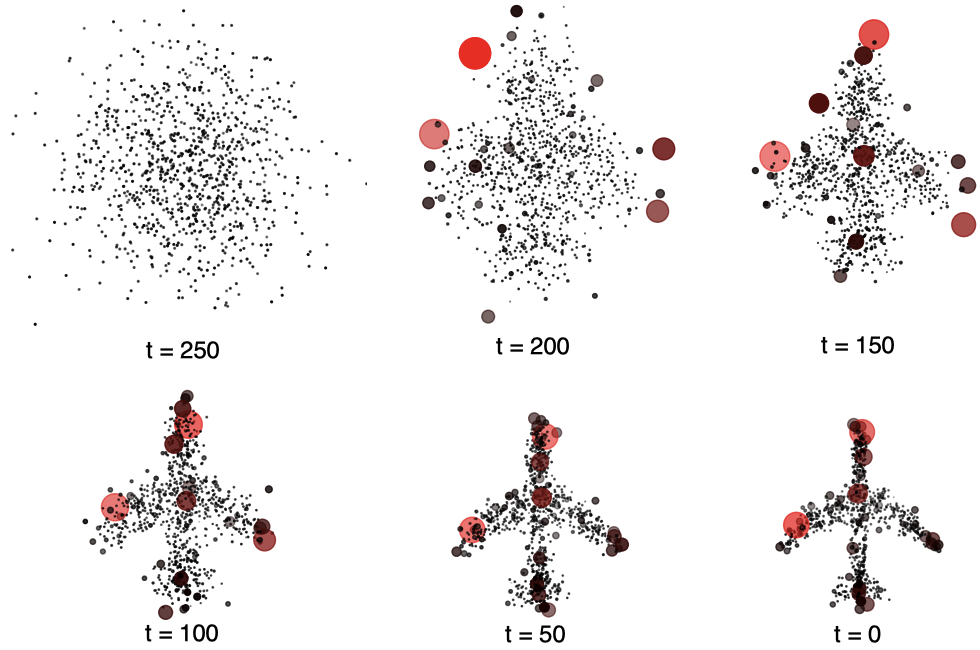}
    \caption{Saliency maps for diffusion process. We integrate the gradients every $50$ steps and calculate the attributions with Integrated Gradients. The redder and larger the points, the more positive the attributions in the prediction. More IGD saliency maps can be found in Fig. \ref{More_IGD}.}
    \label{Fig:CriticalTracking}
    \end{centering}
\end{figure}

In this section we illustrate the results of IGD. In the total number of 250 diffusion steps, we integrate the gradients every 50 steps and calculate the corresponding saliency maps according to Algorithm \ref{algo:tracking}, and randomly choose an example from class ``Airplane" to be illustrated in Fig. \ref{Fig:CriticalTracking}.

The saliency map reveals that the sparse nature \cite{gupta20203d, tan2023explainability} of the point cloud attributions is already formed at the beginning of the reverse diffusion process, and those critical points are also identified at an early stage and their attributions are almost invariant. Moreover, we observe that the critical features within a category are analogous. In Fig. \ref{IGD_Similarity} we complement four additional AM examples for class ``Airplane" and the corresponding diffusion processes. Interestingly, those critical points with the greatest attribution appear only at the tips of noses, wings, and tail, while the points in the centre of the fuselage exhibit relatively smaller attributions. Note that the four examples are generated from models (including the DDPM model, the classifier and its noised version) trained on two different datasets (ModelNet40 and ShapeNet), which indicates the classifier learns similar features from different data source that contribute most to the predictions. Analogues are found for inter-category but similar geometries, as analysed in Sec. \ref{sup:moreIGD}.

For the validity of IGD, we quantitatively evaluate the performance from two aspects, faithfulness and coherence. 

\textbf{Faithfulness}, also known as sensitivity, is one of the most important metrics for explanations. The theory behind faithfulness evaluation is that the confidence of the model prediction decreases dramatically after ablating those features with most positive attributions, and vice versa. In our experiments, We conduct MoRF and LeRF test \cite{bach2015pixel,samek2016evaluating}. For each generated saliency map $\psi$, we recursively ablate $5\%$ of the points with the highest attributions $(x_t, \psi^p_{.05}), (x_t, \psi^p_{.10})\cdots, (x_t, \psi^p_{j})$ and the lowest attributions $(x_t, \psi^n_{.05}), (x_t, \psi^n_{.10})\cdots, (x_t, \psi^n_{j})$ ($j$ is the maximum ablation rate), respectively. We then predict the confidences of these ablated inputs individually with $F$, i.e. $F(x_t/(x_t, \psi^p_{.05})),\cdots,F(x_t/(x_t, \psi^p_{j}))$ and $F(x_t/(x_t, \psi^n_{.05})),\cdots,F(x_t/(x_t, \psi^n_{j}))$. We evaluate the faithfulness of the saliency maps by measuring the areas between the two confidence sequences:

\begin{equation}
    S^j_{F}=\int_{i=0}^{j}F(\psi^p_{i})-F(\psi^n_{i})
\end{equation}

\textbf{Coherence} is a novel metric proposed specifically for explanations in diffusion processes. Recall the DDPM sampling process, where the introduced noise from $x_{t}$ to $x_{t+1}$ is minor, enabling the parameterization of the neural networks for backward diffusion \cite{ho2020denoising}, which indicates that $x_{t+1}$ has significant distributional similarity to $x_t$, with the exception of a small amount of noise. Meanwhile, existing researches \cite{alvarez2018robustness,hancox2020robustness,hsieh2020evaluations} suggest that explainability methods should perform robustly, i.e., the explanations generated for neighboring inputs are supposed to be analogous. Thus, for the explanations of diffusion processes, we leverage numerical new metrics, i.e., the discrepancy between $\psi_{t_1}$ and $\psi_{t_2}$ is negligible when $t_1$ is approaching $t_2$. Quantitatively, we assess two aspects, global stability, which evaluates the statistical robustness of the attributions throughout the diffusion process, and local continuity, which computes the coherence of the explanations for two adjacent sampled $t$. 

For global stability, we compute the variance of all input attributions over the diffusion process: $L_{var}=\frac{\sum_{t=T}^0 var(\psi_t)}{j}$. We assess local continuity with three metrics: the Difference of Predecessor (DF) and the Sliding-Window Average (SWA), which focus on the smoothness of the numerical values, and the Spearman Coefficient Average, which emphasizes the consistency of the rankings. The DF is simply the average of all differences between temporally neighboring attributions $L_{D}=\frac{\sum_{t=T}^{1}\left |\psi_t-\psi_{t-1}  \right |}{T}$. For SWA, we calculate the average of the attributions for three consecutive diffusion samples $\overline{W_t}=\frac{\psi_{t-1}, \psi_{t},\psi_{t+1}}{3}$ (When $t=T$ or $0$, $\psi_{t+1}$ and $\psi_{t-1}$ are ignored, respectively). SWA ($L_W$) is the mean of the difference between all $\psi_t$ and $\overline{W_t}$: $L_{W}=\frac{\sum_{t=T}^{0}\psi_t-\overline{W_t}}{T}$. Similarly, for Spearman Coefficient Average, we compute the mean of the Spearman Coefficients of all attributions with their predecessors $L_{SC}=\frac{\sum_{t=T}^{1}SC(\psi_t,\psi_{t-1})}{T}$, where $SC(a,b)$ denotes the Spearman's Coefficient between $a$ and $b$.

Tab. \ref{tab:IGD_compare} demonstrates the results of the quantitative comparison of IG and IGD. In terms of faithfulness, IGD significantly outperforms IG for both $50\%$ (j = 0.5) and $100\%$ (j = 1.0) ablation, which verifies that integrating the gradients along the diffusion path is more faithful to the model. For coherence, IGD is remarkably more robust, owing to that IGD simply requires additional gradient integration from $x_t$ to $x_{t-1}$, whereas IG recomputes the linear integration from the baseline $x_T$ to $x_{t-1}$, which disrupts the continuity of neighboring samples in diffusion processes.

Interestingly, IG is consistent with the Spearman's coefficients, almost identifying the critical points at $x_T$ period ($L_{SC}=0.986$). However, numerically IG does not exhibit a corresponding consistency ($L_D$, $L_{var}$ and $L_W$), as the vast majority of attributions are centralized to a minority of points, which is in line with the conclusion from existing studies \cite{gupta20203d,tan2023explainability}. We argue that such a sparse attribution may be biased as there exists an alternative integration path that yields saliency maps with significantly higher faithfulness than the linear path from the typical IG.

\section{Conclusion} \label{sec:conclusion}
This work is the first attempt to employ DDPM for generating high-quality AM global explanations, which provides better perceptibility and representativeness than existing methods. In addition, we propose a diffusion path-based attribution approach, which alleviates the bias of the typical Integrated Gradients. In the future work, we would attempt more intuitive visualization approaches for the purpose of shedding light on the intrinsic mechanisms of the point cloud models.
\clearpage
{
    \small
    \bibliographystyle{IEEEtran}
    \bibliography{main}

\begin{thebibliography}{10}
\providecommand{\url}[1]{#1}
\csname url@samestyle\endcsname
\providecommand{\newblock}{\relax}
\providecommand{\bibinfo}[2]{#2}
\providecommand{\BIBentrySTDinterwordspacing}{\spaceskip=0pt\relax}
\providecommand{\BIBentryALTinterwordstretchfactor}{4}
\providecommand{\BIBentryALTinterwordspacing}{\spaceskip=\fontdimen2\font plus
\BIBentryALTinterwordstretchfactor\fontdimen3\font minus \fontdimen4\font\relax}
\providecommand{\BIBforeignlanguage}[2]{{%
\expandafter\ifx\csname l@#1\endcsname\relax
\typeout{** WARNING: IEEEtran.bst: No hyphenation pattern has been}%
\typeout{** loaded for the language `#1'. Using the pattern for}%
\typeout{** the default language instead.}%
\else
\language=\csname l@#1\endcsname
\fi
#2}}
\providecommand{\BIBdecl}{\relax}
\BIBdecl

\bibitem{burkart2021survey}
N.~Burkart and M.~F. Huber, ``A survey on the explainability of supervised machine learning,'' \emph{Journal of Artificial Intelligence Research}, vol.~70, pp. 245--317, 2021.

\bibitem{das2020opportunities}
A.~Das and P.~Rad, ``Opportunities and challenges in explainable artificial intelligence (xai): A survey,'' \emph{arXiv preprint arXiv:2006.11371}, 2020.

\bibitem{van2022explainable}
B.~H. Van~der Velden, H.~J. Kuijf, K.~G. Gilhuijs, and M.~A. Viergever, ``Explainable artificial intelligence (xai) in deep learning-based medical image analysis,'' \emph{Medical Image Analysis}, p. 102470, 2022.

\bibitem{nazir2023survey}
S.~Nazir, D.~M. Dickson, and M.~U. Akram, ``Survey of explainable artificial intelligence techniques for biomedical imaging with deep neural networks,'' \emph{Computers in Biology and Medicine}, p. 106668, 2023.

\bibitem{cambria2023survey}
E.~Cambria, L.~Malandri, F.~Mercorio, M.~Mezzanzanica, and N.~Nobani, ``A survey on xai and natural language explanations,'' \emph{Information Processing \& Management}, vol.~60, no.~1, p. 103111, 2023.

\bibitem{danilevsky2020survey}
M.~Danilevsky, K.~Qian, R.~Aharonov, Y.~Katsis, B.~Kawas, and P.~Sen, ``A survey of the state of explainable ai for natural language processing,'' \emph{arXiv preprint arXiv:2010.00711}, 2020.

\bibitem{pomerleau2015review}
F.~Pomerleau, F.~Colas, R.~Siegwart \emph{et~al.}, ``A review of point cloud registration algorithms for mobile robotics,'' \emph{Foundations and Trends{\textregistered} in Robotics}, vol.~4, no.~1, pp. 1--104, 2015.

\bibitem{cheng2020morphing}
Q.~Cheng, P.~Sun, C.~Yang, Y.~Yang, and P.~X. Liu, ``A morphing-based 3d point cloud reconstruction framework for medical image processing,'' \emph{Computer methods and programs in biomedicine}, vol. 193, p. 105495, 2020.

\bibitem{cui2021deep}
Y.~Cui, R.~Chen, W.~Chu, L.~Chen, D.~Tian, Y.~Li, and D.~Cao, ``Deep learning for image and point cloud fusion in autonomous driving: A review,'' \emph{IEEE Transactions on Intelligent Transportation Systems}, vol.~23, no.~2, pp. 722--739, 2021.

\bibitem{tan2023visualizing}
H.~Tan, ``Visualizing global explanations of point cloud dnns,'' in \emph{Proceedings of the IEEE/CVF Winter Conference on Applications of Computer Vision}, 2023, pp. 4741--4750.

\bibitem{qi2017pointnet}
C.~R. Qi, H.~Su, K.~Mo, and L.~J. Guibas, ``Pointnet: Deep learning on point sets for 3d classification and segmentation,'' in \emph{Proceedings of the IEEE conference on computer vision and pattern recognition}, 2017, pp. 652--660.

\bibitem{qi2017pointnet++}
C.~R. Qi, L.~Yi, H.~Su, and L.~J. Guibas, ``Pointnet++: Deep hierarchical feature learning on point sets in a metric space,'' \emph{Advances in neural information processing systems}, vol.~30, 2017.

\bibitem{erhan2009visualizing}
D.~Erhan, Y.~Bengio, A.~Courville, and P.~Vincent, ``Visualizing higher-layer features of a deep network,'' \emph{University of Montreal}, vol. 1341, no.~3, p.~1, 2009.

\bibitem{sundararajan2017axiomatic}
M.~Sundararajan, A.~Taly, and Q.~Yan, ``Axiomatic attribution for deep networks,'' in \emph{International conference on machine learning}.\hskip 1em plus 0.5em minus 0.4em\relax PMLR, 2017, pp. 3319--3328.

\bibitem{li2020learning}
J.~Li, V.~Nagarajan, G.~Plumb, and A.~Talwalkar, ``A learning theoretic perspective on local explainability,'' \emph{arXiv preprint arXiv:2011.01205}, 2020.

\bibitem{simonyan2013deep}
K.~Simonyan, A.~Vedaldi, and A.~Zisserman, ``Deep inside convolutional networks: Visualising image classification models and saliency maps,'' \emph{arXiv preprint arXiv:1312.6034}, 2013.

\bibitem{bach2015pixel}
S.~Bach, A.~Binder, G.~Montavon, F.~Klauschen, K.-R. M{\"u}ller, and W.~Samek, ``On pixel-wise explanations for non-linear classifier decisions by layer-wise relevance propagation,'' \emph{PloS one}, vol.~10, no.~7, p. e0130140, 2015.

\bibitem{selvaraju2017grad}
R.~R. Selvaraju, M.~Cogswell, A.~Das, R.~Vedantam, D.~Parikh, and D.~Batra, ``Grad-cam: Visual explanations from deep networks via gradient-based localization,'' in \emph{Proceedings of the IEEE international conference on computer vision}, 2017, pp. 618--626.

\bibitem{springenberg2014striving}
J.~T. Springenberg, A.~Dosovitskiy, T.~Brox, and M.~Riedmiller, ``Striving for simplicity: The all convolutional net,'' \emph{arXiv preprint arXiv:1412.6806}, 2014.

\bibitem{ribeiro2016should}
M.~T. Ribeiro, S.~Singh, and C.~Guestrin, ``" why should i trust you?" explaining the predictions of any classifier,'' in \emph{Proceedings of the 22nd ACM SIGKDD international conference on knowledge discovery and data mining}, 2016, pp. 1135--1144.

\bibitem{lundberg2017unified}
S.~M. Lundberg and S.-I. Lee, ``A unified approach to interpreting model predictions,'' \emph{Advances in neural information processing systems}, vol.~30, 2017.

\bibitem{petsiuk2018rise}
V.~Petsiuk, A.~Das, and K.~Saenko, ``Rise: Randomized input sampling for explanation of black-box models,'' \emph{arXiv preprint arXiv:1806.07421}, 2018.

\bibitem{verma2020counterfactual}
S.~Verma, V.~Boonsanong, M.~Hoang, K.~E. Hines, J.~P. Dickerson, and C.~Shah, ``Counterfactual explanations and algorithmic recourses for machine learning: A review,'' \emph{arXiv preprint arXiv:2010.10596}, 2020.

\bibitem{mothilal2020explaining}
R.~K. Mothilal, A.~Sharma, and C.~Tan, ``Explaining machine learning classifiers through diverse counterfactual explanations,'' in \emph{Proceedings of the 2020 conference on fairness, accountability, and transparency}, 2020, pp. 607--617.

\bibitem{goyal2019counterfactual}
Y.~Goyal, Z.~Wu, J.~Ernst, D.~Batra, D.~Parikh, and S.~Lee, ``Counterfactual visual explanations,'' in \emph{International Conference on Machine Learning}.\hskip 1em plus 0.5em minus 0.4em\relax PMLR, 2019, pp. 2376--2384.

\bibitem{wachter2017counterfactual}
S.~Wachter, B.~Mittelstadt, and C.~Russell, ``Counterfactual explanations without opening the black box: Automated decisions and the gdpr,'' \emph{Harv. JL \& Tech.}, vol.~31, p. 841, 2017.

\bibitem{sushil2018rule}
M.~Sushil, S.~{\v{S}}uster, and W.~Daelemans, ``Rule induction for global explanation of trained models,'' \emph{arXiv preprint arXiv:1808.09744}, 2018.

\bibitem{hailesilassie2016rule}
T.~Hailesilassie, ``Rule extraction algorithm for deep neural networks: A review,'' \emph{arXiv preprint arXiv:1610.05267}, 2016.

\bibitem{nguyen2015deep}
A.~Nguyen, J.~Yosinski, and J.~Clune, ``Deep neural networks are easily fooled: High confidence predictions for unrecognizable images,'' in \emph{Proceedings of the IEEE conference on computer vision and pattern recognition}, 2015, pp. 427--436.

\bibitem{nguyen2016synthesizing}
A.~Nguyen, A.~Dosovitskiy, J.~Yosinski, T.~Brox, and J.~Clune, ``Synthesizing the preferred inputs for neurons in neural networks via deep generator networks,'' \emph{Advances in neural information processing systems}, vol.~29, 2016.

\bibitem{nguyen2017plug}
A.~Nguyen, J.~Clune, Y.~Bengio, A.~Dosovitskiy, and J.~Yosinski, ``Plug \& play generative networks: Conditional iterative generation of images in latent space,'' in \emph{Proceedings of the IEEE conference on computer vision and pattern recognition}, 2017, pp. 4467--4477.

\bibitem{mahendran2016visualizing}
A.~Mahendran and A.~Vedaldi, ``Visualizing deep convolutional neural networks using natural pre-images,'' \emph{International Journal of Computer Vision}, vol. 120, pp. 233--255, 2016.

\bibitem{yosinski2015understanding}
J.~Yosinski, J.~Clune, A.~Nguyen, T.~Fuchs, and H.~Lipson, ``Understanding neural networks through deep visualization,'' \emph{arXiv preprint arXiv:1506.06579}, 2015.

\bibitem{mordvintsev2015inceptionism}
A.~Mordvintsev, C.~Olah, and M.~Tyka, ``Inceptionism: Going deeper into neural networks,'' 2015.

\bibitem{zhou2017activation}
Z.~Zhou, H.~Cai, S.~Rong, Y.~Song, K.~Ren, W.~Zhang, Y.~Yu, and J.~Wang, ``Activation maximization generative adversarial nets,'' \emph{arXiv preprint arXiv:1703.02000}, 2017.

\bibitem{lan2019modeling}
S.~Lan, R.~Yu, G.~Yu, and L.~S. Davis, ``Modeling local geometric structure of 3d point clouds using geo-cnn,'' in \emph{Proceedings of the IEEE/cvf conference on computer vision and pattern recognition}, 2019, pp. 998--1008.

\bibitem{liu2019relation}
Y.~Liu, B.~Fan, S.~Xiang, and C.~Pan, ``Relation-shape convolutional neural network for point cloud analysis,'' in \emph{Proceedings of the IEEE/CVF conference on computer vision and pattern recognition}, 2019, pp. 8895--8904.

\bibitem{komarichev2019cnn}
A.~Komarichev, Z.~Zhong, and J.~Hua, ``A-cnn: Annularly convolutional neural networks on point clouds,'' in \emph{Proceedings of the IEEE/CVF conference on computer vision and pattern recognition}, 2019, pp. 7421--7430.

\bibitem{riegler2017octnet}
G.~Riegler, A.~Osman~Ulusoy, and A.~Geiger, ``Octnet: Learning deep 3d representations at high resolutions,'' in \emph{Proceedings of the IEEE conference on computer vision and pattern recognition}, 2017, pp. 3577--3586.

\bibitem{klokov2017escape}
R.~Klokov and V.~Lempitsky, ``Escape from cells: Deep kd-networks for the recognition of 3d point cloud models,'' in \emph{Proceedings of the IEEE international conference on computer vision}, 2017, pp. 863--872.

\bibitem{zeng20183dcontextnet}
W.~Zeng and T.~Gevers, ``3dcontextnet: Kd tree guided hierarchical learning of point clouds using local and global contextual cues,'' in \emph{Proceedings of the European Conference on Computer Vision (ECCV) Workshops}, 2018, pp. 0--0.

\bibitem{simonovsky2017dynamic}
M.~Simonovsky and N.~Komodakis, ``Dynamic edge-conditioned filters in convolutional neural networks on graphs,'' in \emph{Proceedings of the IEEE conference on computer vision and pattern recognition}, 2017, pp. 3693--3702.

\bibitem{wang2019dynamic}
Y.~Wang, Y.~Sun, Z.~Liu, S.~E. Sarma, M.~M. Bronstein, and J.~M. Solomon, ``Dynamic graph cnn for learning on point clouds,'' \emph{Acm Transactions On Graphics (tog)}, vol.~38, no.~5, pp. 1--12, 2019.

\bibitem{chen2019clusternet}
C.~Chen, G.~Li, R.~Xu, T.~Chen, M.~Wang, and L.~Lin, ``Clusternet: Deep hierarchical cluster network with rigorously rotation-invariant representation for point cloud analysis,'' in \emph{Proceedings of the IEEE/CVF conference on computer vision and pattern recognition}, 2019, pp. 4994--5002.

\bibitem{zheng2019pointcloud}
T.~Zheng, C.~Chen, J.~Yuan, B.~Li, and K.~Ren, ``Pointcloud saliency maps,'' in \emph{Proceedings of the IEEE/CVF International Conference on Computer Vision}, 2019, pp. 1598--1606.

\bibitem{gupta20203d}
A.~Gupta, S.~Watson, and H.~Yin, ``3d point cloud feature explanations using gradient-based methods,'' in \emph{2020 International Joint Conference on Neural Networks (IJCNN)}.\hskip 1em plus 0.5em minus 0.4em\relax IEEE, 2020, pp. 1--8.

\bibitem{tan2022surrogate}
H.~Tan and H.~Kotthaus, ``Surrogate model-based explainability methods for point cloud nns,'' in \emph{Proceedings of the IEEE/CVF Winter Conference on Applications of Computer Vision}, 2022, pp. 2239--2248.

\bibitem{dhariwal2021diffusion}
P.~Dhariwal and A.~Nichol, ``Diffusion models beat gans on image synthesis,'' \emph{Advances in Neural Information Processing Systems}, vol.~34, pp. 8780--8794, 2021.

\bibitem{sohl2015deep}
J.~Sohl-Dickstein, E.~Weiss, N.~Maheswaranathan, and S.~Ganguli, ``Deep unsupervised learning using nonequilibrium thermodynamics,'' in \emph{International Conference on Machine Learning}.\hskip 1em plus 0.5em minus 0.4em\relax PMLR, 2015, pp. 2256--2265.

\bibitem{ho2020denoising}
J.~Ho, A.~Jain, and P.~Abbeel, ``Denoising diffusion probabilistic models,'' \emph{Advances in Neural Information Processing Systems}, vol.~33, pp. 6840--6851, 2020.

\bibitem{Luo_2021_CVPR}
S.~Luo and W.~Hu, ``Diffusion probabilistic models for 3d point cloud generation,'' in \emph{Proceedings of the IEEE/CVF Conference on Computer Vision and Pattern Recognition (CVPR)}, June 2021, pp. 2837--2845.

\bibitem{lyu2021conditional}
Z.~Lyu, Z.~Kong, X.~Xu, L.~Pan, and D.~Lin, ``A conditional point diffusion-refinement paradigm for 3d point cloud completion,'' \emph{arXiv preprint arXiv:2112.03530}, 2021.

\bibitem{nichol2022point}
A.~Nichol, H.~Jun, P.~Dhariwal, P.~Mishkin, and M.~Chen, ``Point-e: A system for generating 3d point clouds from complex prompts,'' \emph{arXiv preprint arXiv:2212.08751}, 2022.

\bibitem{li2022diffusionpointlabel}
T.~Li, Y.~Fu, X.~Han, H.~Liang, J.~J. Zhang, and J.~Chang, ``Diffusionpointlabel: Annotated point cloud generation with diffusion model,'' in \emph{Computer Graphics Forum}, vol.~41, no.~7.\hskip 1em plus 0.5em minus 0.4em\relax Wiley Online Library, 2022, pp. 131--139.

\bibitem{nichol2021improved}
A.~Q. Nichol and P.~Dhariwal, ``Improved denoising diffusion probabilistic models,'' in \emph{International Conference on Machine Learning}.\hskip 1em plus 0.5em minus 0.4em\relax PMLR, 2021, pp. 8162--8171.

\bibitem{sturmfels2020visualizing}
P.~Sturmfels, S.~Lundberg, and S.-I. Lee, ``Visualizing the impact of feature attribution baselines,'' \emph{Distill}, 2020, https://distill.pub/2020/attribution-baselines.

\bibitem{wu20153d}
Z.~Wu, S.~Song, A.~Khosla, F.~Yu, L.~Zhang, X.~Tang, and J.~Xiao, ``3d shapenets: A deep representation for volumetric shapes,'' in \emph{Proceedings of the IEEE conference on computer vision and pattern recognition}, 2015, pp. 1912--1920.

\bibitem{ma2022rethinking}
X.~Ma, C.~Qin, H.~You, H.~Ran, and Y.~Fu, ``Rethinking network design and local geometry in point cloud: A simple residual mlp framework,'' \emph{arXiv preprint arXiv:2202.07123}, 2022.

\bibitem{tan2023explainability}
H.~Tan and H.~Kotthaus, ``Explainability-aware one point attack for point cloud neural networks,'' in \emph{Proceedings of the IEEE/CVF Winter Conference on Applications of Computer Vision}, 2023, pp. 4581--4590.

\bibitem{samek2016evaluating}
W.~Samek, A.~Binder, G.~Montavon, S.~Lapuschkin, and K.-R. M{\"u}ller, ``Evaluating the visualization of what a deep neural network has learned,'' \emph{IEEE transactions on neural networks and learning systems}, vol.~28, no.~11, pp. 2660--2673, 2016.

\bibitem{alvarez2018robustness}
D.~Alvarez-Melis and T.~S. Jaakkola, ``On the robustness of interpretability methods,'' \emph{arXiv preprint arXiv:1806.08049}, 2018.

\bibitem{hancox2020robustness}
L.~Hancox-Li, ``Robustness in machine learning explanations: Does it matter?'' in \emph{Proceedings of the 2020 conference on fairness, accountability, and transparency}, 2020, pp. 640--647.

\bibitem{hsieh2020evaluations}
C.-Y. Hsieh, C.-K. Yeh, X.~Liu, P.~Ravikumar, S.~Kim, S.~Kumar, and C.-J. Hsieh, ``Evaluations and methods for explanation through robustness analysis,'' \emph{arXiv preprint arXiv:2006.00442}, 2020.

\end{thebibliography}
}
\clearpage
\section{Supplementary} \label{sec:supplementary}
\beginsupplement
\setcounter{section}{6}
\subsection{Technical details and parameter configuration} \label{sec:traindetail}
In this section we elaborate on the parameter configurations of each model while training.

$\mathbf{F}$: $F$ is the classifier to be explained, and thus the only requirement is to achieve as satisfactory an accuracy as possible. We followed the implementation of \cite{qi2017pointnet} to train a classical PointNet on ModelNet40, which achieves a test accuracy of $89.2\%$.

$\mathbf{F'}$: $F'$ is the classifier trained with the noised dataset to guide the gradients in the early stages of the diffusion process. There are two differences in $F'$ compared to $F$: the addition of noise and the introduction of $t$. For the former, in order to approximate the training data more closely to the product of diffusion, we directly generate them with the trained $p_{\theta}$ and $t$ following equation \ref{equa:diffusion}. For $t$, to avoid exploding values, we encode decimal with binary: $De(t)\rightarrow V_{Bi(t)}$. However, the number of digits in binary encoding is variable which cannot be implemented in neural networks. The length of vector reserved for the encoding of $t$ are:
\begin{equation}
    len(V_{Bi(t)})=\left \lceil  \log_2(T)\right \rceil
\end{equation}
In our experiment, $T=250$, hence $len(V_{Bi(t)}) = 8$. There are numerous options for where $V_{Bi(t)}$ can be embedded, such as concatenating on the global features after Max-pooling. We plug it in directly after the coordinates of each point. During training, we randomly sample $t$ for each batch to obtain the noise-added $q(x_t|x_0)$, which is used as an input to predict the label l. The final test accuracy of the trained F' is $81.2\%$.

$\mathbf{p_{\theta}}$: We set $T=250$ as a balance between generating quality and speed. The scheduling of $beta$ is from $1e-4$ to $2e-2$, the guidance scale $s$ is $1e-4$. During training, the learning rate progressively decays from $lr=1e-2$ to $1e-4$. In addition, we set the number of attention headers for both the encoder and decoder of the PDT to 3 and the widths are 64, 128, and 256, respectively.

\subsection{Backgrounds of the evaluation metrics}\label{sup:intrometrics}

For quantitative evaluation of AM-based global explanations three main indicators are considered, namely representativeness, diversity and perceptibility. Representativeness refers to the extent to which the explanations activate the target neuron, perceptibility denotes the extent to which the generated explanations can be recognized by humans, and diversity indicates the extent to which the profiles of the explanations distinguish themselves from each other. In order to comprehensively evaluate point cloud global explanations, we follow the metric proposed by \cite{tan2023visualizing}: PCAMS. 

PCAMS is formulated as

\begin{equation}
\begin{aligned}
    PCAMS = & IS_m(x_g) \\
    - & \frac{(log(FID(x_g,x_i)) + log(CD(x_g,x_i)))}{2}
\end{aligned}
\end{equation}

$\mathbf{IS_m}$ is the Modified Inception Score (m-IS), which assesses the representativeness and diversity ($IS_m=e^{{\mathbb{E}_{x_i}}[\mathbb{E}_{x_j}[(\mathbb{KL}(p(y|x_i)||p(y|x_j))]]}$). 

$\mathbf{CD}$ denotes Chamfer Distance, which measures the point-wise distance between point cloud instances. We utilize CD to measure the point-wise distance between the generated explanations and samples randomly sampled from real data as the perceptibility evaluation metric. $CD$ is formulated as:

\begin{equation}
    CD(x_g,x_i)=\frac{1}{|x_g|}\sum_{p_m\in x_g}\min_{p_n\in x_i}\left \| p_m-p_n \right \|_2
\end{equation}
where $x_g$ and $x_i$ are the generated global explanations and sampled real data, respectively.

$\mathbf{FID}$ indicates Fréchet Inception Distances, which assesses the latent distance between two point cloud instances. Instead of point-wise discrepancies, FID evaluates the perceptual similarity of two inputs, which approximates more to human recognition mechanisms. FID is formulated as:

\begin{equation}
    FID_{PN}=\left \| \mu_i -\mu_g  \right \|^{2}+Tr(\sigma_i + \sigma_g - 2(\sigma_i \sigma_g)^{\frac{1}{2}})
\end{equation}
where $\mu$ and $\sigma$ are the mean and variance of the latent vectors, respectively, and the subscripts $g$ and $i$ are the generated explanations and randomly sampled real data, respectively. Practically, we input $x_g$ and $x_i$ into PointNet respectively, and intercept the two outputs at the middle layer of the neural network (after the pooling layer and before FC3, respectively) and compute their FID distances.

$\mathbf{EMD}$ denotes Earth Mover's Distance, which is another point-wise distance measurement. EMD focuses more on the perception of overall density than CD. Therefore, EMD is less affected by outliers, at the expense of less attention to details between the inputs. EMD is formulated as:

\begin{equation}
EMD(x_g,x_i)=\frac{\sum_{j=1}^{m}\sum_{k=1}^{n}Pr_{j,k}d_{j,k}}{\sum_{j=1}^{m}\sum_{k=1}^{n}Pr_{j,k}}
\end{equation}
where $Pr$ represents a pair of points chosen from $x_g$ and $x_i$. In PCAMS, EMD is not involved in the computation, but we employ it in a separate evaluation as an alternative reference metric for point-wise distances.

\subsection{More DAM visualizations} \label{sup:MoreDAMVisu}
\begin{figure*}
    \begin{centering}
    \includegraphics[width=0.65\textwidth]{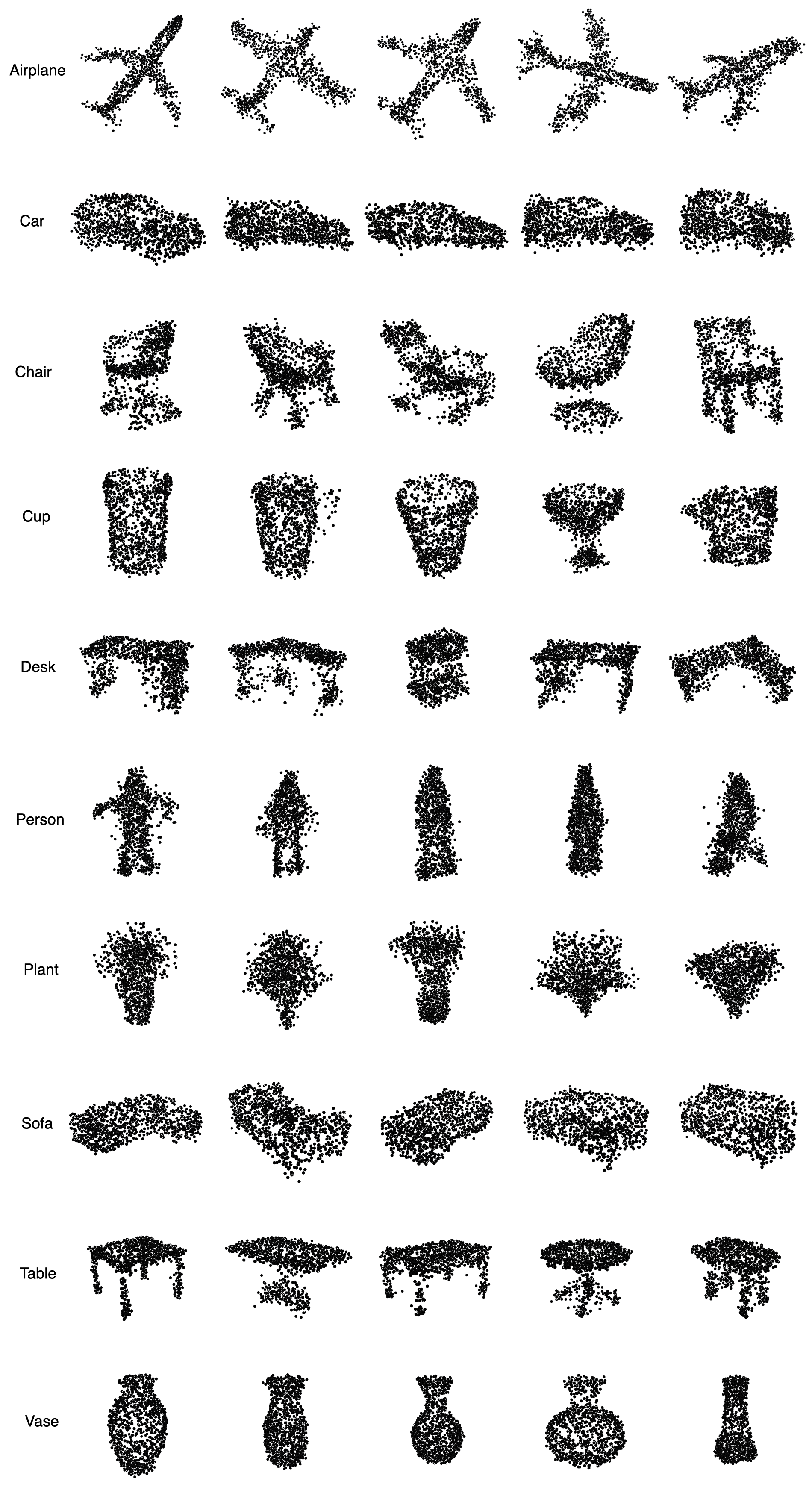}
    \caption{More global explanations generated by DAM.}
    \label{More_DAM}
    \end{centering}
\end{figure*}

This section we demonstrate more global explanations generated with DAM. We select 10 common ones from the 40 categories of ModelNet40 and utilize DAM to generate 5 global explanations for them respectively. Fig. \ref{More_DAM} illustrates the generated global explanations. It can be seen that the explanations generated by DAM possess both perceptibility and diversity.

\subsection{More IGD visualizations} \label{sup:moreIGD}
\begin{figure*}
    \begin{centering}
    \includegraphics[width=0.72\textwidth]{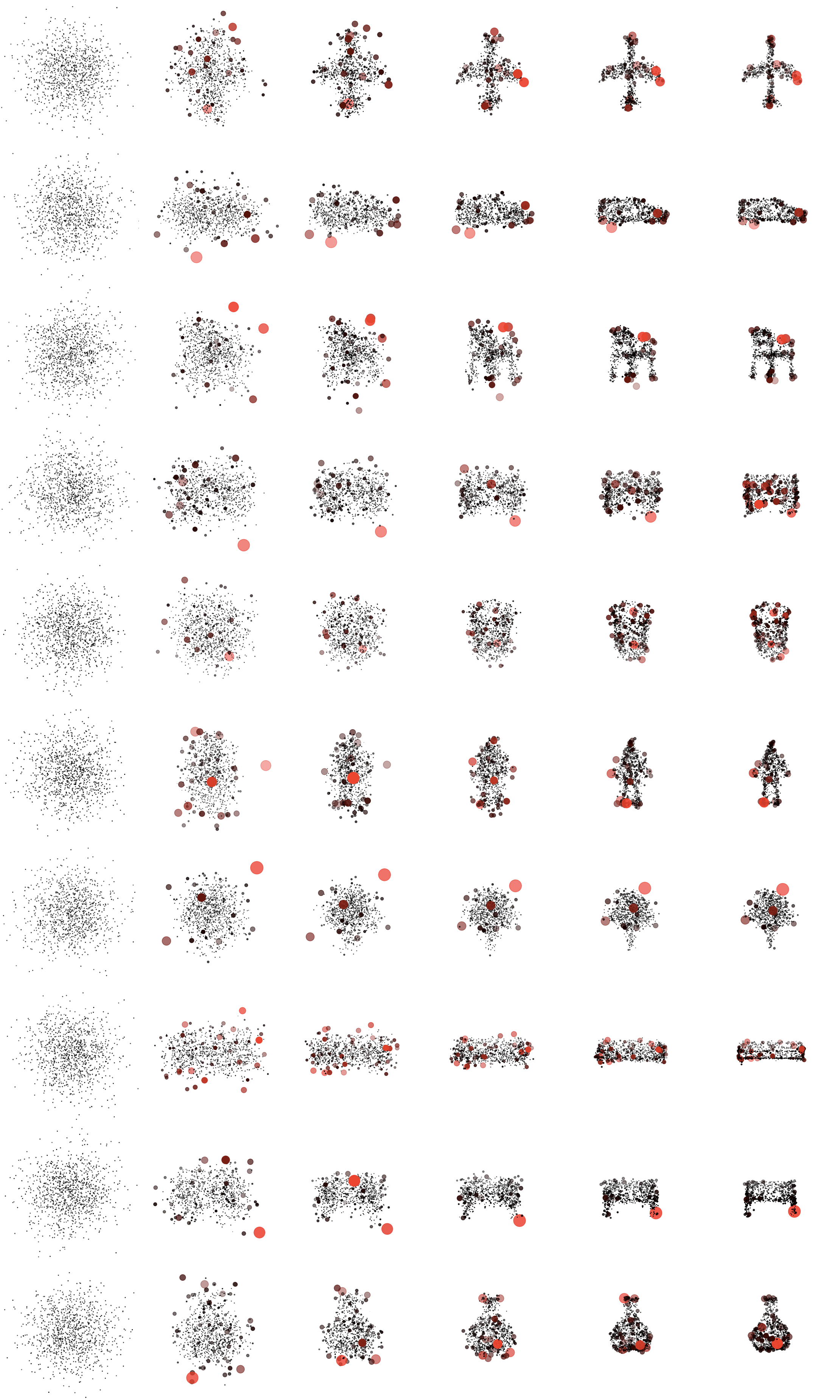}
    \caption{More examples of IGD visualizations. The classes demonstrated are, from top to bottom: Airplane, Car, Chair, Desk, Cup, Person, Plant, Sofa, Table, Vase, respectively. From left to right: $t=250$, $200$. $150$, $100$, $50$, $0$, respectively}
    \label{More_IGD}
    \end{centering}
\end{figure*}

In this section we show additional saliency maps generated by IGD. As shown in Fig. \ref{More_IGD}, we select 10 common classes from ModelNet40 and explain their diffusion process with IGD.

\begin{figure*}
    \begin{centering}
    \includegraphics[width=1.0\textwidth]{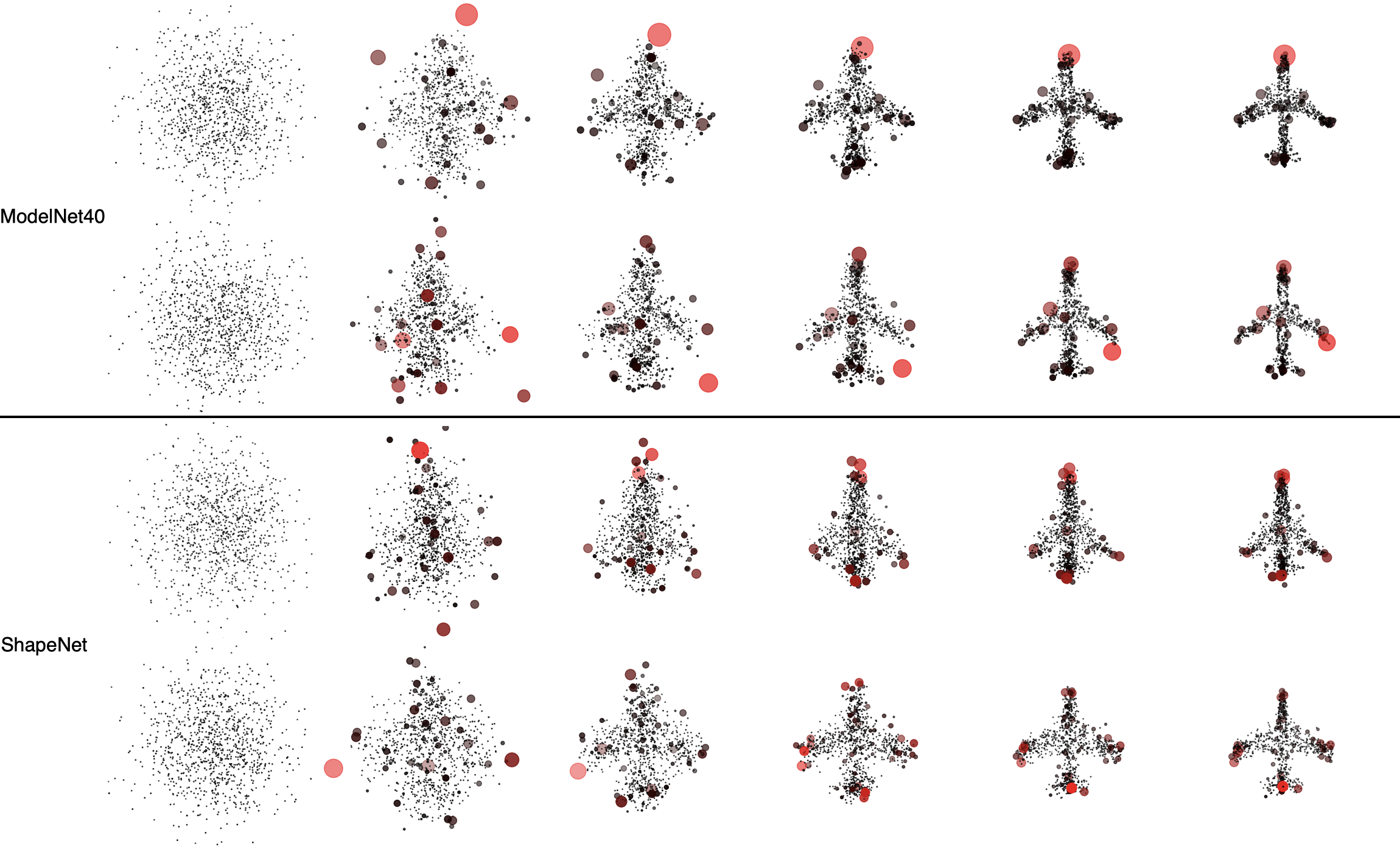}
    \caption{Diffusion process of four examples from class ``Airplane" and their IGD saliency maps. The two examples above and below are from ModelNet40 and ShapeNet, respectively.}
    \label{IGD_Similarity}
    \end{centering}
\end{figure*}

In order to visualise the similarity of IGD for the same category, we show four saliency maps for the class "Airplane" in Figure \ref{IGD_Similarity}. It can be observed that critical features occur consistently at specific positions, such as noses, fuselages and wing tips, which indicates that the model predictions are based on specific point cloud structures.

\begin{figure*}
    \begin{centering}
    \includegraphics[width=1.0\textwidth]{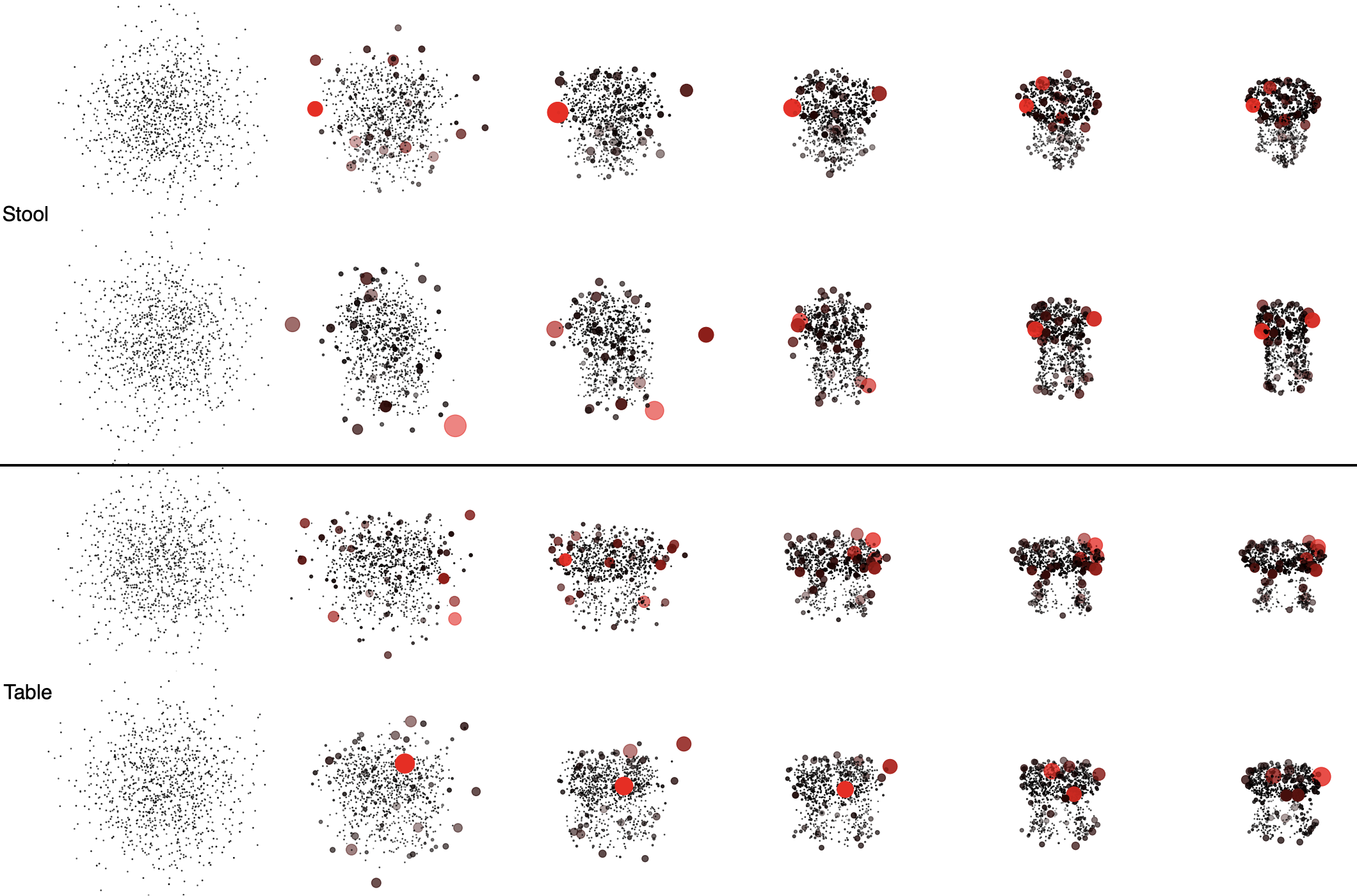}
    \caption{Diffusion process and IGD of four examples from class ``Stool" (upper part) and ``Table" (lower part), respectively. }
    \label{tab:Stool_Table}
    \end{centering}
\end{figure*}

For further validation, we present AM examples from two different classes with similar geometries, i.e., stools and tables (round tables), in Fig. \ref{tab:Stool_Table}. Even though they do not belong to the same category, both of them share a similar structure, i.e., a circular plane. It can be noticed that the model consistently focuses on the edges of the circular platform.

\subsection{Detailed structure of the PDT} \label{sup:structure_PDT}

\begin{figure*}
    \begin{centering}
    \includegraphics[width=0.8\textwidth]{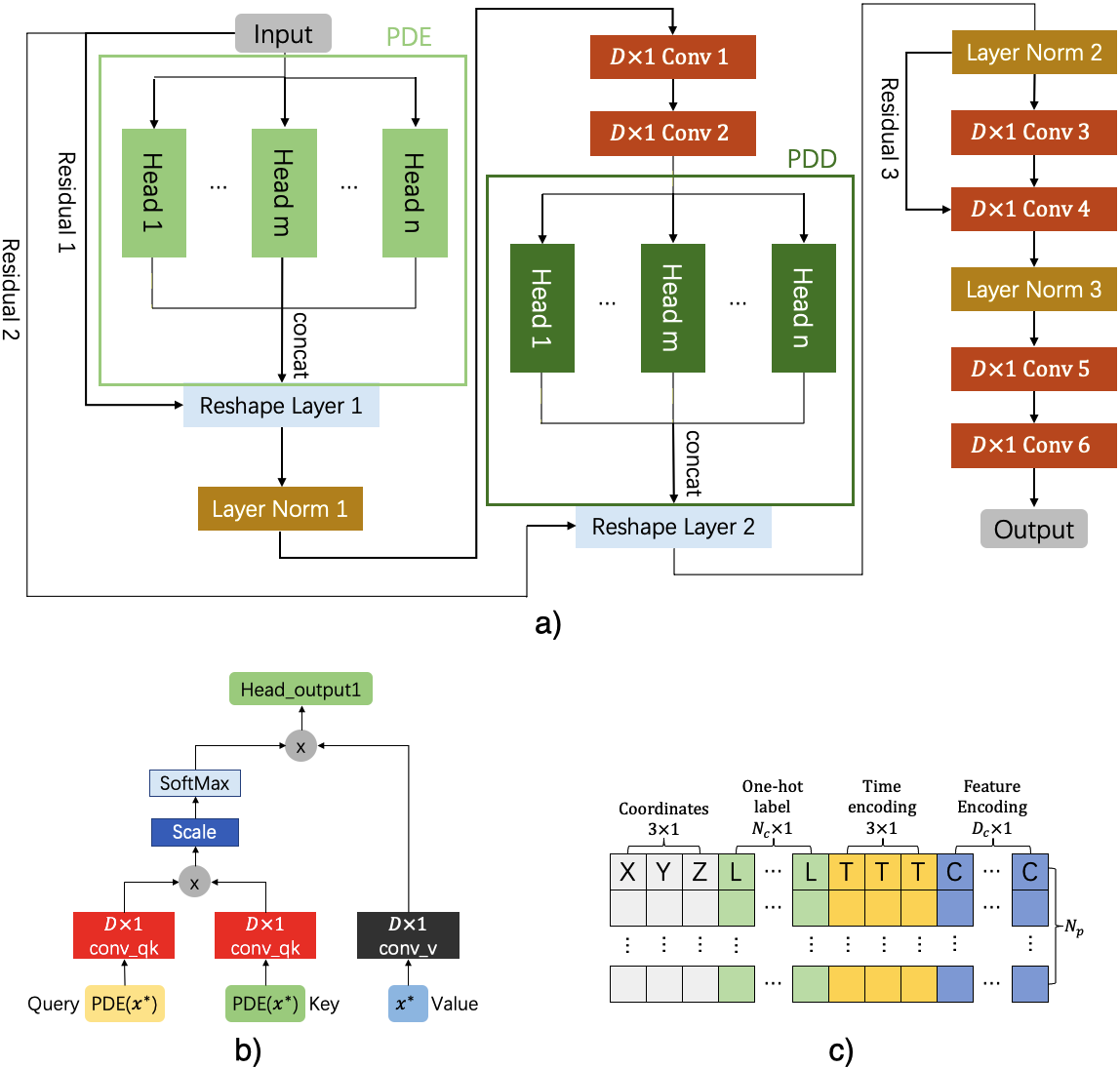}
    \caption{a) Detailed architecture of the Point Diffusion Transformer. It consists mainly of a Point Diffusion Encoder (PDE) and a Point Diffusion Decoder (PDD), which includes an adjustable number of attention headers. b) Internal structure of a PDD attention head. c) Components of point cloud input vectors, each point carries feature vectors such as time and label.}
    \label{PDT_general}
    \end{centering}
\end{figure*}

In this section we illustrate the detailed structure of PDT. Fig. \ref{PDT_general} presents the architecture of PDT, which consists of two major components: Point Diffusion Encoder (PDE) and Point Diffusion Decoder (PDD). Both PDE and PDD incorporate an adjustable number of attention heads. Notably, the entire PDT is completely symmetric in structure, i.e., the output is independent of the order of the points, while the points are independent and neighboring points do not interfere with each other. The input global information is spliced after the coordinates of the points by transforming them into $1 \times d$ vectors, thus also guaranteeing the integrity of the global and local information.

\subsection{Visualizing non-output layers with DAM} \label{sup:visu_other_layers}

\begin{figure*}
    \begin{centering}
    \includegraphics[width=0.6\textwidth]{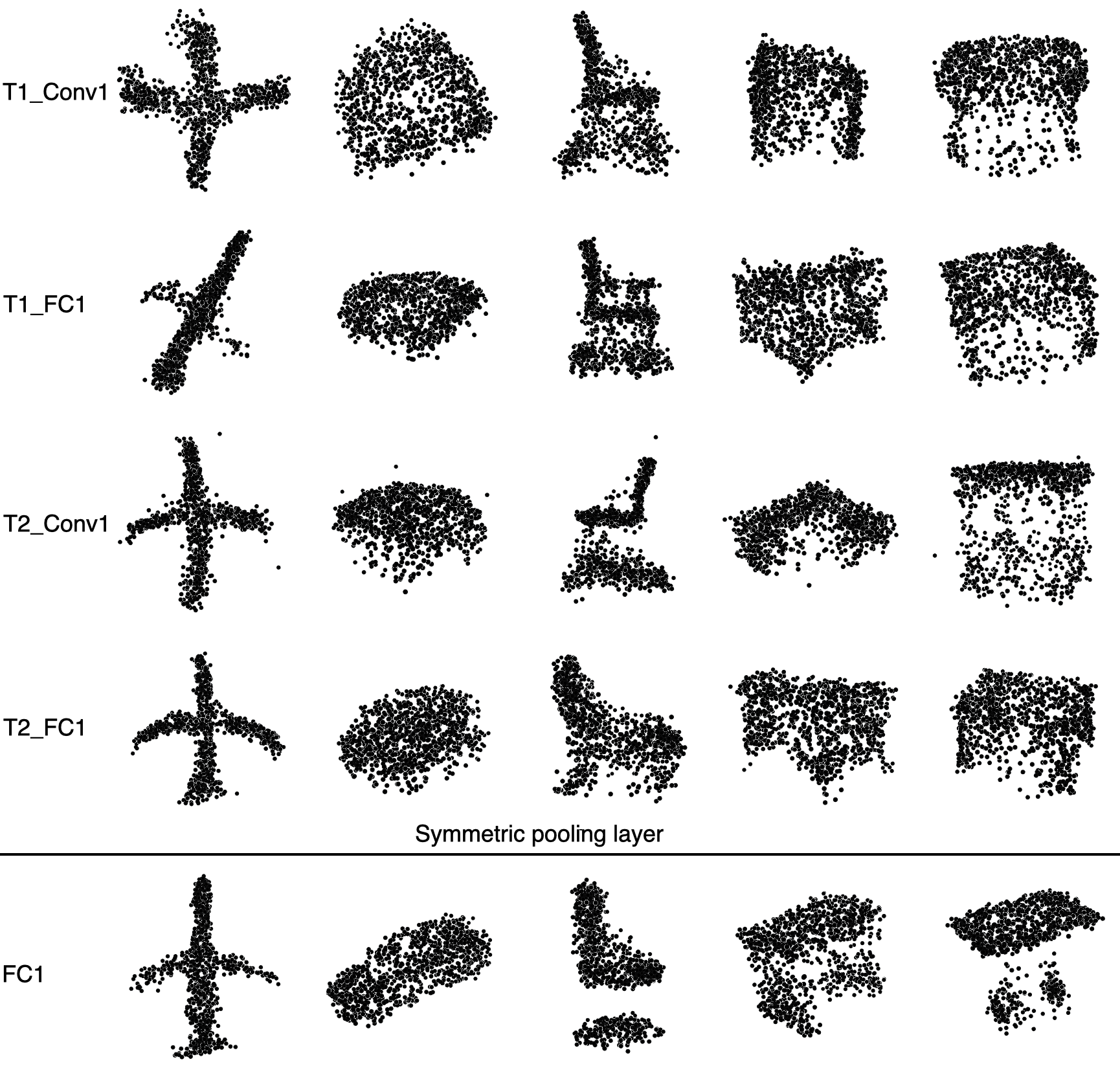}
    \caption{DAM visualization of other layers of PointNet. From top to bottom are: the first convolutional layer and its first fully connected layer for the first and T-network, the first convolutional layer and its first fully connected layer for the second T-network, and the first fully connected layer for the final global features, where the first four layers are before the global symmetry function (i.e., Max-pooling) and can be treated as local features, and the last layer is after the global symmetry function and is treated as global features.}
    \label{view_mult_layer}
    \end{centering}
\end{figure*}

In addition to the final activation layer, we also visualize the intermediate layers of the network. We select the first convolutional layer and the first fully connected layer of the 2 T-Nets for visualization, respectively. These structures are ahead of the global pooling layer and can therefore be considered as point-wise local feature extraction. In addition, we chose the fully connected layer after the pooling, where point-wise features have been transformed into global ones. Fig. \ref{view_mult_layer} illustrates the results of visualizing these intermediate layers of PointNet with DAM. Interestingly, what DAM visualizes from these low-level layers is more of a rough outline of the object and contains a lot of outliers, such as airplanes, cars, and chairs. The results of DAM for higher-level visualizations exhibit objects with more complete structures and fewer outliers. This is intuitive as the lower layers of the point cloud network only pay attention to point-wise (or local) features, and global information is not included in these neurons, whereas after global pooling, the entire structure of the object is incorporated into the higher-level neurons.

\subsection{Visualizing multiple neurons with DAM} \label{sup:two_neurons}

\begin{figure}
    \begin{centering}
    \includegraphics[width=0.475\textwidth]{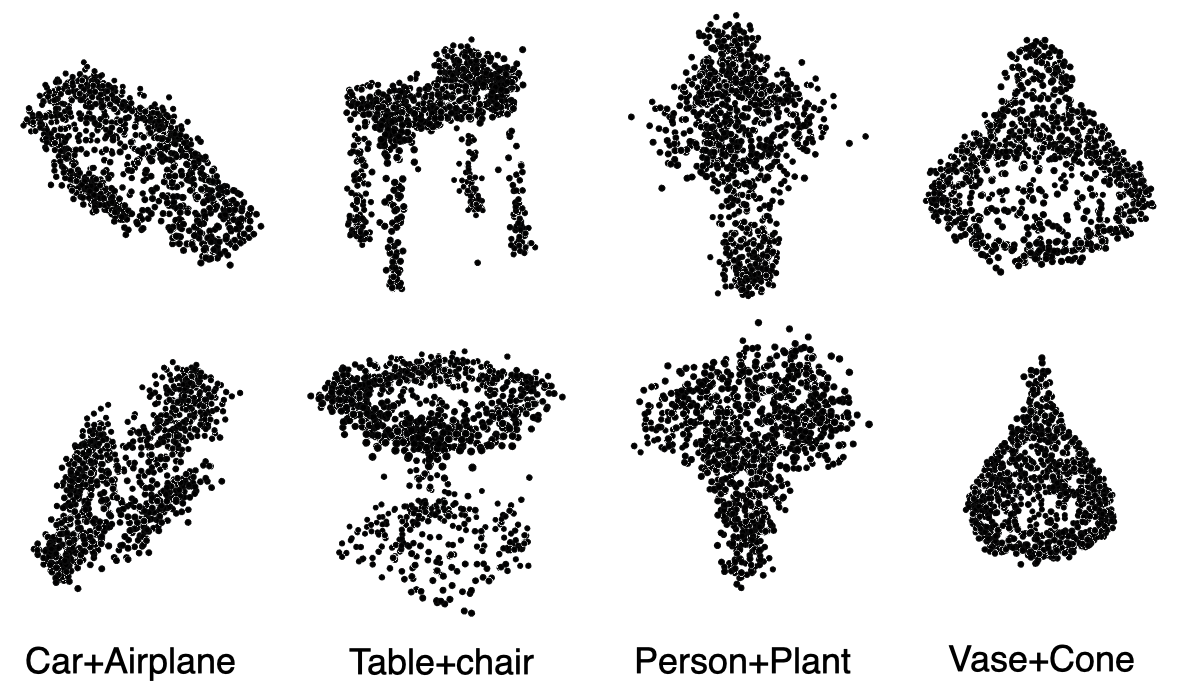}
    \caption{Examples of visualizations that maximize two activations at the same time with DAM.}
    \label{AM_double_neuron}
    \end{centering}
\end{figure}

We perform another visualization test on DAM: maximizing two activations simultaneously and generating perceptible instances via PDT. However, PDT is a model that is label dependent, i.e., the label $l_1$ of the visualization target needs to be included in the input. When two activations need to be maximized in parallel, PDT cannot receive both labels at the same time. As a solution, we leverage alternating iterations: the inputs of PDT and the guide gradients from the classifiers F (F') target $l_1$ when $t$ is even, and $l_2$ if $t$ is odd. Fig. \ref{AM_double_neuron} illustrates the visualization of the maximization for multiple activations. It can be observed that when optimizing the gradient to two labels at the same time, DAM fuses the features of two objects together, e.g., a flat car body with wings extended on both sides, a table with a backrest, and a cone with a bottle top. This provides side evidence of which features are representative for the corresponding activation.

\subsection{Feature-wise independence} \label{sup:featindependent}
To verify the superiority of feature-independent models, we train PointwiseNet \cite{Luo_2021_CVPR}, another network containing fully connected layers, as a reference. These fully-connected layers do not possess ``symmetry" in the diffusion process of the point cloud, and thus the correlation between points is not completely eliminated in PointwiseNet. We train two versions of PointwiseNet, the original one, i.e., proposed in \cite{Luo_2021_CVPR}, and the targeted version, i.e., with labels incorporated as guidance. Table \ref{tab:PWNvsPDT} presents the quantitative evaluations of the generated global explanations. The original non-targeted PointwiseNet is incapable of generating high-quality global explanations, and when label guidance is introduced, the generation quality is significantly improved. However, there is no remarkable growth in m-IS, suggesting that the AM iterative process is hampered. Although PDT suffers from a negligible gap in explanation quality compared to the targeted PointwiseNet, it has a noticeable superiority in terms of representativeness and diversity (m-IS).

\begin{table*}[]
\centering
\begin{tabular}{cccccccc}
\hline
Property                        & Model                     & mSR $\uparrow$   & m-IS $\uparrow$ & FID $\downarrow$ & CD $\downarrow$ & EMD $\downarrow$ & PCAMS $\uparrow$ \\ \hline
\multirow{2}{*}{Points-related} & PointWiseNet              & 3.8\%           & 1.412           & 0.014            & 0.069           & 177.34           & 4.88             \\
                                & PointWiseNet (label)      & \textbf{70.2\%} & 1.438           & \textbf{0.008}   & \textbf{0.044}  & \textbf{126.77}  & 5.41             \\ \hline
Points-independent              & PointDiffusionTransformer & 61.5\%          & \textbf{1.781}  & 0.009            & 0.045  & 133.97           & \textbf{5.68}    \\ \hline
\end{tabular}
\caption{Comparison of quantitative evaluations generated by feature-independent PDT and feature-related PointwiseNet. A new metric is introduced here: mSR, the success rate, representing the rate at which a generated explanation is correctly classified by the classifier $F$ into the class it should belong to. Since DAM can only be performed in the sampling phase of the diffusion process, there is an upper limit to the number of its iterations. Non-optimal structures or modules may obstruct the gradient guidance, resulting in the generated explanations failing the prediction test of the classifier $F$.} \label{tab:PWNvsPDT}
\end{table*}

\subsection{Randomize $x$ vs. $z$} \label{sec_random_x_z}

In the beginning of diffusion sampling, we consider the following two initialization methods:

\begin{itemize}
    \item $\mathbf{z\rightarrow x_{T}}$: In the method proposed by \cite{Luo_2021_CVPR}, they directly randomize the reparameterized vector $z$, which reduces the computation intensity without compromising the generative performance of the diffusion. 
    
    \item $\mathbf{x_r\rightarrow z\rightarrow x_{T}}$: Randomizing the reparameterized $z$ may result in generating samples that $F$ has never seen before, thus confounding the gradient guidance. Compared to the first method, we incorporate two extra steps, i.e., randomize $x_r$ and encode it by $q_{\varphi}(z|x_0)$, and then reparameterize to $z$. 
\end{itemize}

\begin{figure*}
    \begin{centering}
    \includegraphics[width=0.7\textwidth]{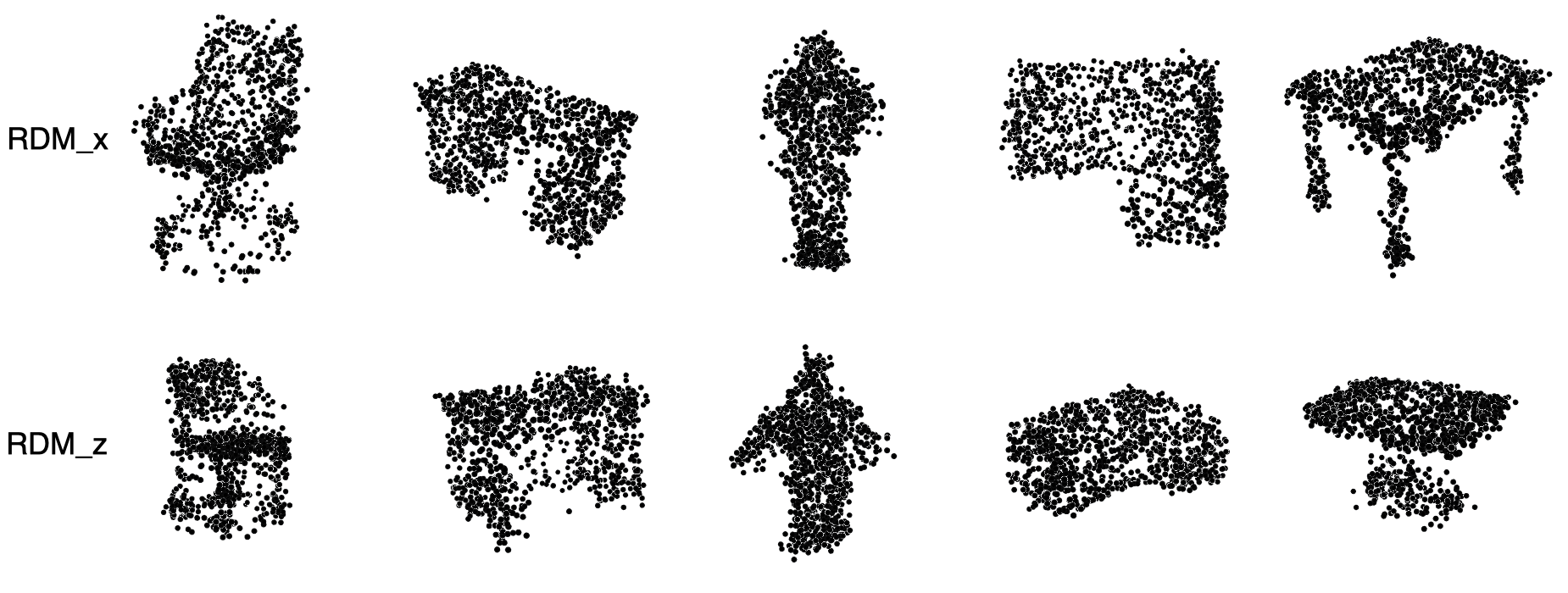}
    \caption{Two different initialization randomization methods. The top and bottom rows are randomizing $x$ and randomizing $z$, respectively.}
    \label{fig:ini_compare}
    \end{centering}
\end{figure*}

We qualitatively and quantitatively evaluate both initialization approaches and demonstrate the results in Tab. \ref{tab:randomxvsz} and Fig. \ref{fig:ini_compare}. It can be seen that there is insignificant difference between the two initialization methods in terms of perceptibility of the generated explanations, which is also reflected in the quantitative metrics $FID$, $CD$ and $EMD$. However, our approach dramatically enhances representativeness, i.e., the degree to which target activation is maximized, as demonstrated by $m-IS$. Therefore, our initialization method improves the final score of AM samples.

\begin{table*}[]
\centering
\begin{tabular}{clccccc}
\hline
       & mSR $\uparrow$           & m-IS $\uparrow$ & FID $\downarrow$ & CD $\downarrow$ & EMD $\downarrow$ & PCAMS $\uparrow$ \\ \hline
RDM\_z & \textbf{68.0\%} & 1.38            & 0.009            & \textbf{0.043}  & \textbf{122.51}  & 5.31             \\
RDM\_x & 61.5\%          & \textbf{1.78}   & \textbf{0.009}   & 0.045           & 133.97           & \textbf{5.68}    \\
\hline
\end{tabular}
\caption{Performance comparison of randomly initializing $x$ vs. $z$.}\label{tab:randomxvsz}
\end{table*}

\subsection{AM on Logits vs. SoftMax} \label{sec:logitsvssoftmax}

When employing AM algorithms, there are multiple activation options, including 1) Activation is the neuron on the logits layer.2) Activation is the neuron on the SoftMax layer.3) Activation is the neuron on the logits of the SoftMax layer. Existing research \cite{nguyen2017plug} suggests that the global explanations generated in the third case perform the best as the gradient of the input $x$ by its logits layer $l$ ($\frac{\partial l_i}{\partial x}$) is independent from the multiplier from the softmax layer $s_i$. We verify this conclusion on DAM, as shown in Tab. \ref{tab:logitsvssoftmax}. While the choice of logits and SoftMax as activations achieve excellent success rates for AM generation, their performances are significantly inferior compared to log(SoftMax). In particular, on the $m-IS$ metric, logits and SoftMax activations barely exhibit representativeness and diversity (note that the minimum score for $m-IS$ is $1.0$).

\begin{figure}
    \begin{centering}
    \includegraphics[width=0.475\textwidth]{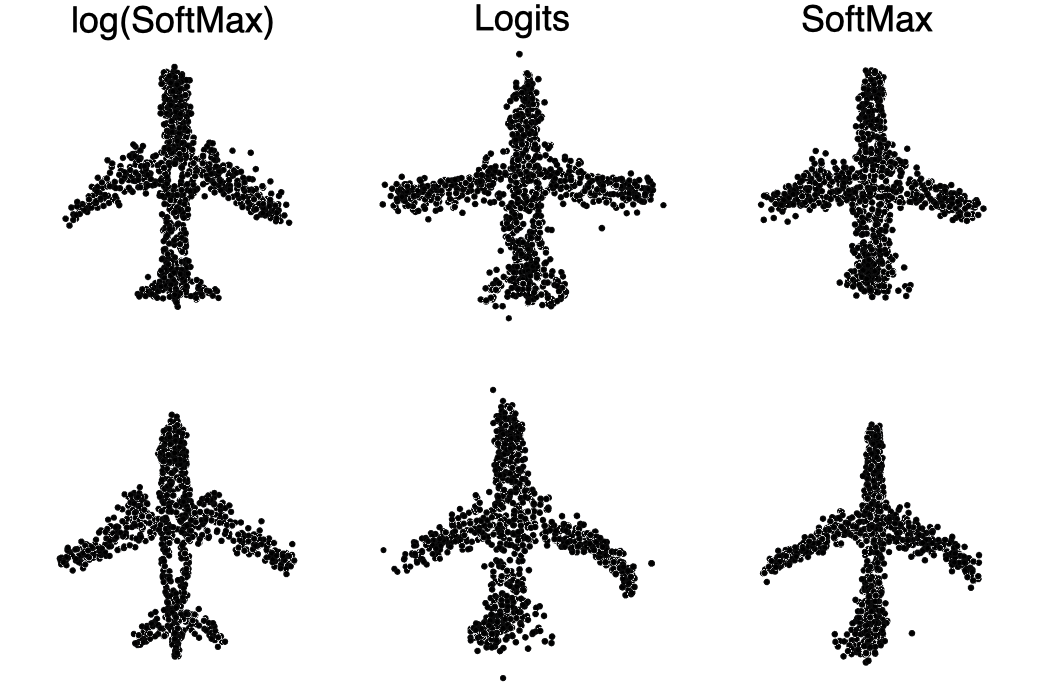}
    \caption{Visual comparison of global explanations generated by the three different activation choices.}
    \label{Fig:logitsvsoftmax}
    \end{centering}
\end{figure}

For reference, Fig. \ref{Fig:logitsvsoftmax} illustrates two explanations of class ``Airplane" for each activation choice. Obviously, explanations generated by logit activation suffer from more outliers, while those generated by SoftMax activation are structurally malformed.

\begin{table*}[]
\centering
\begin{tabular}{ccccccc}
\hline
             & mSR $\uparrow$ & m-IS $\uparrow$ & FID $\downarrow$ & CD $\downarrow$ & EMD $\downarrow$ & PCAMS $\uparrow$ \\ \hline
Logits       & \textbf{99.0\%} & 1.049           & 0.049            & 0.054           & 134.69           & 4.01             \\
SoftMax      & 99.7\%          & 1.134           & 0.011            & 0.055           & 142.26           & 4.83             \\
log(SoftMax) & 61.5\%          & \textbf{1.781}  & \textbf{0.009}   & \textbf{0.045}  & \textbf{133.97}  & \textbf{5.68}    \\ \hline
\end{tabular}
\caption{Quantitative evaluation of global explanations generated by three different activation choices.}\label{tab:logitsvssoftmax}
\end{table*}

\subsection{Input for attention headers of PDD} \label{sec:qkvstudy}

In this section we discuss the impact of the inputs to the attention head of the decoder on the generation of the final global explanations.

The inputs to the attention head of the decoder that can be considered include the encoded input $PDE(x)$ (or with residual $PDE(x) + x$), the original input $x$ and their combinations. In PDT, the inputs of Query, Key and Value of the attention heads in PointDiffusionDecoder (PDD) are $PDE(x)$, $PDE(x)$ and the original input $x$, respectively. Experiments demonstrate that this combination achieves a better balance between AM success rate and explanation perceptibility. In Tab. \ref{tab:qkvstudy}, we demonstrate four additional combinations: a) all $PDE(x)$, b) all $x$ except Query which is $PDE(x)$, c) all $PDE(x)$ except Query which is $x$ and d) all $x$ except Value which is $PDE(x)$. We train each combination for 200k iterations, and then generate the corresponding global explanations in conjunction with DAM and evaluate them quantitatively. We observe that the incorporation of $x$ in Value significantly improves the AM success rate and dramatically enhances the representativeness and diversity of the explanations (compared with the first row). Furthermore, if Key is set to $x$ at the same time, the AM success rate drops to a lower level, although with a slight boost in mIS (the second row). If $x$ is taken as the input to Query, although it raises the success rate to nearly $100\%$, the perceptibility of the generated explanations diminishes significantly, and they are barely representative and diverse (the third and fourth rows). 

\begin{figure*}
    \begin{centering}
    \includegraphics[width=0.8\textwidth]{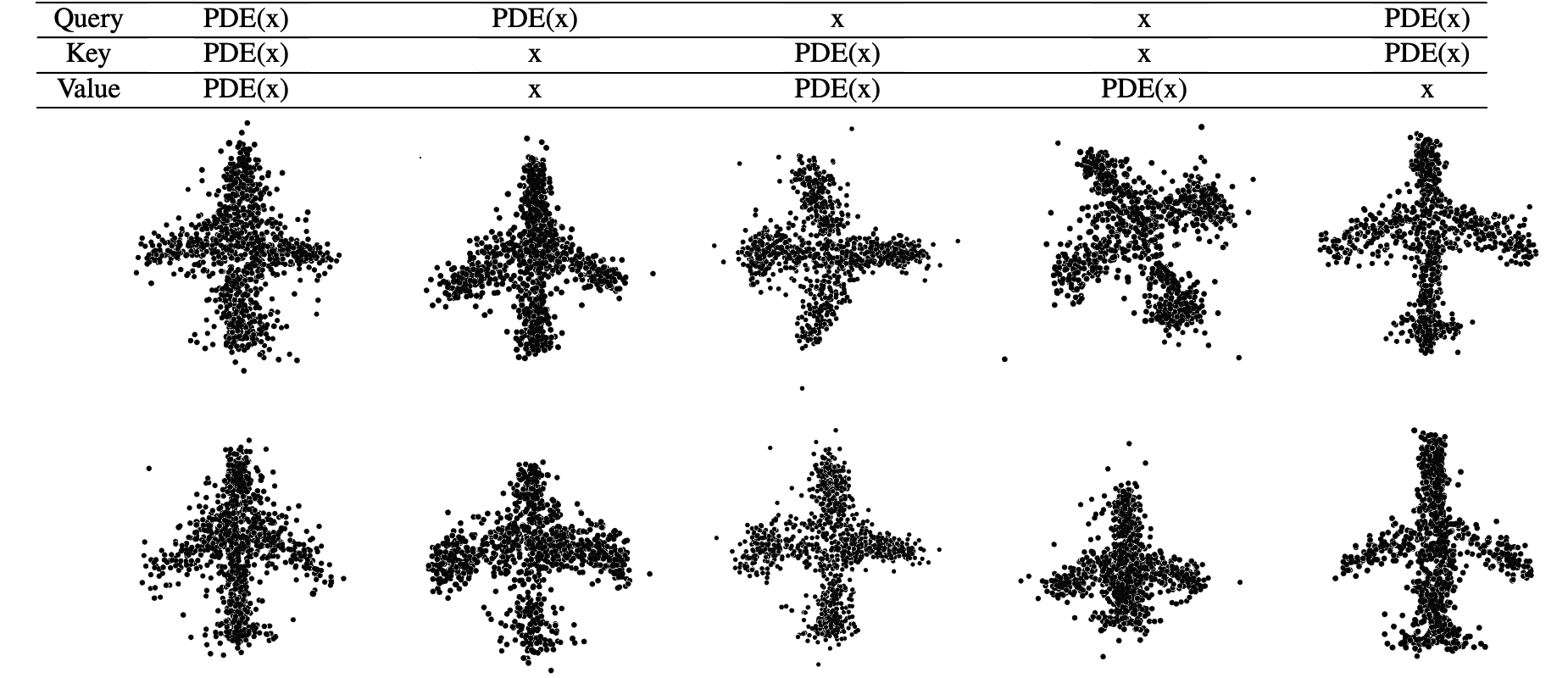}
    \caption{Two examples of explanations generated from different combinations of attention header inputs. Where $En(x)$ represents the latent features that are encoded by PointDiffusionEncoder and $x$ represents the original input.}
    \label{fig:qkvstudy}
    \end{centering}
\end{figure*}

For intuition, we present in Fig. \ref{fig:qkvstudy} two examples of explanations generated by each input combination, with the category ``Airplane". It can be seen that when $PDE(x)$ is the input to the Query and the Key and Value are identical, the generated explanation has a complete outline that can be perceived by humans. When the input to Query is x, the explanations suffer from more outliers and the structure is distorted. The explanations generated by PDT (the last column) possess both structural perceptibility and at the same time fewer outliers.

\begin{table*}[]
\centering
\begin{tabular}{ccc|cccccc}
\hline
Query & Key   & Value & mSR$\uparrow$    & m-IS $\uparrow$ & FID $\downarrow$ & CD $\downarrow$ & EMD $\downarrow$ & PCAMS $\uparrow$ \\ \hline
PDE(x) & PDE(x) & PDE(x) & 8.75\%           & 1.206           & 0.011            & 0.042           & 127.28           & 5.03             \\
PDE(x) & x     & x     & 6.25\%           & \textbf{1.599}  & \textbf{0.009}   & \textbf{0.036}  & 121.73           & \textbf{5.61}    \\
x     & PDE(x) & PDE(x) & \textbf{95.00\%} & 1.161           & 0.038            & 0.045           & 137.94           & 4.33             \\
x     & x     & PDE(x) & 89.25\%          & 1.360           & 0.030            & 0.056           & 143.66           & 4.54             \\
PDE(x) & PDE(x) & x     & 17.75\%          & 1.532           & \textbf{0.009}   & 0.042           & \textbf{121.04}  & 5.47             \\ \hline
\end{tabular}
\caption{Quantitative evaluation of different global explanations generated by various combinations of attention header inputs.}\label{tab:qkvstudy}
\end{table*}

\subsection{Dual Guide VS Single Guide} \label{subsection:eval_noised}
As the inputs approximate Gaussian distributions at the beginning of the diffusion process, while the classifier to be explained $F$ is trained on well-shaped instances, the guiding gradients may be deceptive to AM generation. Training another noisy version of the classifier $F'$ dramatically enhances the explaination performance, although it prolongs the processing time. Tab. \ref{tab:noisedvsunnoised} presents the guiding effect of $F'$. It can be seen that the promotion of incorporating $F'$ is mainly reflected in the m-IS, i.e., the gradient guide of $F$ is smoothed by adopting $F'$ as a transition. 

\begin{table*}[]
\centering
\begin{tabular}{cccccccc}
\hline
        & mAcc.    & mSR $\uparrow$              & m-IS $\uparrow$  & FID $\downarrow$ & CD $\downarrow$  & EMD $\downarrow$  & PCAMS $\uparrow$ \\ \hline
w/o. F' & $89.2\%$ & $\mathbf{63.75\%}$ & $1.374$          & $0.012$          & $\mathbf{0.045}$ & $\mathbf{132.49}$          & $5.11$          \\ \hline
w. F'   & $81.3\%$ & $61.5\%$           & $\mathbf{1.781}$ & $\mathbf{0.009}$ & $\mathbf{0.045}$          & $133.97$ & $\mathbf{5.68}$  \\ \hline
\end{tabular}
\caption{Quantitative comparison of the performance on generated explanations with the inclusion of the noise-trained classifier $F'$ versus its exclusion. mAcc. is the prediction accuracy on the test set at the end of training.}
\label{tab:noisedvsunnoised}
\end{table*}

\end{document}